%% file: ezgikorkmazaaai.tex
\def\year{2022}\relax
\documentclass[letterpaper]{article} 
\usepackage{aaai22}  
\usepackage{times}  
\usepackage{helvet}  
\usepackage{courier}  
\usepackage[hyphens]{url}  
\usepackage{graphicx} 
\usepackage{natbib}  
\usepackage{caption} 
\DeclareCaptionStyle{ruled}{labelfont=normalfont,labelsep=colon,strut=off} 
\usepackage{algorithm}
\usepackage{algorithmic}
\input{math_commands.tex}

\usepackage{stackengine}
\usepackage{subfig}
\usepackage{multicol}
\usepackage{booktabs}

\newcommand{\norm}[1]{\left\lVert#1\right\rVert}
\newcommand{\abs}[1]{\left\lvert#1\right\rvert}
\usepackage{newfloat}
\usepackage{listings}
\floatstyle{ruled}
\newfloat{listing}{tb}{lst}{}
\floatname{listing}{Listing}
\title{Deep Reinforcement Learning Policies Learn Shared Adversarial Features Across MDPs}
\author{
		Ezgi Korkmaz
}

\affiliations{
		ezgikorkmazk@gmail.com

}

\usepackage{bibentry}

\begin{document}

\maketitle

\begin{abstract}
The use of deep neural networks as function approximators has led to striking progress for reinforcement learning algorithms and applications. Yet the knowledge we have on decision boundary geometry and the loss landscape of neural policies is still quite limited. In this paper we propose a framework to investigate the decision boundary and loss landscape similarities across states and across MDPs. We conduct experiments in various games from Arcade Learning Environment, and discover that high sensitivity directions for neural policies are correlated across MDPs. We argue that these high sensitivity directions support the hypothesis that non-robust features are shared across training environments of reinforcement learning agents. We believe our results reveal fundamental properties of the environments used in deep reinforcement learning training, and represent a tangible step towards building robust and reliable deep reinforcement learning agents.
\end{abstract}

\section{Introduction}
\label{submission}

Building on the success of DNNs for image classification, deep reinforcement learning has seen remarkable advances in various complex environments \citet{mn15, schulman17,lil15}. Along with these successes come new challenges stemming from the current lack of understanding of the structure of the decision boundary and loss landscape of neural network policies.
Notably, it has been shown that the high sensitivity of DNN image classifiers to imperceptible perturbations to inputs also occurs for neural policies \citet{huang17, lin17, korkmaz20}. This lack of robustness is especially critical for deep reinforcement learning, where the actions taken by the agent can have serious real-life consequences \citet{levin18}.

Recent work has shown that adversarial examples are a consequence of the existence of non-robust (i.e. adversarial) features of datasets used in image classifier training \citet{ilyas19}. That is, there are certain features which are actually useful in classifying the data, but are extremely sensitive to small perturbations, and thus incomprehensible to humans.

The existence in standard datasets of non-robust features, which both generalize well but are simultaneously highly sensitive to small perturbations, raise serious concerns about the way that DNN models are currently trained and tested. On the one hand, since non-robust features are actually useful in classification, current training methods which optimize for classification accuracy have no reason to ignore them. On the other hand, since these features can be altered with visually imperceptible perturbations they present a formidable obstacle to constructing models that behave anything like humans. Instead, models trained in the presence of non-robust features are likely to have directions of high-sensitivity (or equivalently small margin) correlated with these features.

In the reinforcement learning setting, where policies are generally trained in simulated environments, the presence of non-robust features leading to high-sensitivity directions for neural network policies raises serious concerns about the ability of these policies to generalize beyond their training simulations. Consequently, identifying the presence of non-robust features in simulated training environments and understanding their effects on neural-network policies is an important first step in designing algorithms for training reinforcement learning agents which can perform well in real-world settings.

In this paper we study the decision boundaries of neural network policies in deep reinforcement learning. In particular, we ask: how are high-sensitivity directions related to the actions taken across states for a neural network policy? Are high-sensitivity directions correlated between neural policies trained in different MDPs? Do state-of-the-art adversarially trained deep reinforcement learning policies inherit similar high-sensitivity directions to vanilla trained deep reinforcement learning policies? Do non-robust features exist in the deep reinforcement learning training environments?
To answer these questions, we propose a framework based on identifying directions of low distance to the neural policy decision boundary, and investigating how these directions affect the decisions of the agents across states and MDPs. Our main contributions are as follows:

\begin{itemize}
\item We introduce a framework based on computing a high-sensitivity direction in one state, and probing the decision boundary of the neural policy along this direction as the neural policy is executed in a set of carefully controlled scenarios.
\item We examine the change in the action distribution of an agent whose input is shifted along a high-sensitivity direction, and show that in several cases these directions correspond to shifts towards the same action across different states. These results lend credence to the hypothesis that high-sensitivity directions for neural policies correspond to non-robust discriminative features.
\item We investigate the state-of-the-art adversarially trained deep reinforcement learning policies and show that adversarially trained deep reinforcement learning policies share high sensitivity directions with vanilla trained deep reinforcement learning policies.
\item Via experiments in the Arcade Learning Environment we rigorously show that the high-sensitivity directions computed in our framework correlate strongly across states and in several cases across MDPs. This suggests that distinct MDPs from standard baseline environments contain correlated non-robust features that are utilized by deep neural policies.
\end{itemize}

\section{Related Work and Background}

\subsection{Deep Reinforcement Learning}

In this paper we examine discrete action space MDPs that are represented by a tuple: $\mathcal{M}=(S,A,P,r,\gamma,s_0)$ where $S$ is a set of states, $A$ is a set of discrete actions, $P:S \times A \times S \to \R$ is the transition probability, $r: S \times A \to \R$ is the reward function, $\gamma$ is the discount factor, and $s_0$ is the initial state distribution. The agent interacts with the environment by observing $s \in S$, taking actions $a \in A$, and receiving rewards $r: S \times A \to \R$. A policy $\pi: S \times A \to \R$ assigns a probability distribution over actions $\pi(s,\cdot)$ to each state $s$.
The goal in reinforcement learning is to learn a policy $\pi$ that maximizes the expected cumulative discounted reward $R = \mathbb{E}[\sum_{t=0}^{T-1}\gamma^tr(s_t,a_t)]$ where $a_t \sim \pi(s_t,\cdot)$. For an MDP $\mathcal{M}$ and policy $\pi$ we call a sequence of state, action, reward, next state tuples, $(s_i,a_i,r_i,s_i')$, that occurs when utilizing $\pi$ in $\mathcal{M}$ an episode. We use $p_{\mathcal{M},\pi}$ to denote the probability distribution over the episodes generated by the randomness in $\mathcal{M}$ and the policy $\pi$.
In Q-learning the goal of maximizing the expected discounted cumulative rewards is achieved by building a state-action value function $Q(s,a)= \mathbb{E}_{s \sim \pi(s,\cdot)} \left[\sum_{t=0}^\infty \gamma^t r(s_t,a_t)|s_0=s,a_0=a\right]$. The function $Q(s,a)$ is intended to  represent the expected cumulative rewards obtained by taking action $a$ in state $s$, and in all future states $s'$ taking the action $a'$ which maximizes $Q(s',a')$.

\subsection{Adversarial Perturbation Methods}

In order to identify high-sensitivity directions for neural policies we use methods designed to compute adversarial perturbations of minimal $\ell_p$-norm.
The first paper to discuss adversarial examples for DNNs was \citet{szegedy13}.
Subsequently \citet{fellow15} introduced the fast gradient sign method (FGSM) which computes adversarial perturbations by maximizing the linearization of the cost function in the $\ell_{\infty}$-ball of radius $\epsilon$.
\begin{equation}
\mathnormal{\displaystyle x_{\textrm{adv}} = x+ \epsilon \cdot \frac{\nabla_{x}J(\displaystyle x,y)}{\norm{\nabla_{x}J(x,y)}_p},}
\end{equation}
Here $x$ is the clean input, $y$ is the correct label for $x$, and $J$ is the cost function used to train the neural network classifier.
Further improvements were obtained by \citet{kurakin16} by using FGSM gradients to iteratively search within the $\ell_p$-norm ball of radius $\epsilon$.

Currently, the state-of-the-art method for computing minimal adversarial perturbations is the formulation proposed by \citet{carlini17}. In particular, \citet{anis18} and \citet{tramer20} showed that the C\&W formulation can overcome the majority of the proposed defense and detection algorithms. For this reason, in this work we create the adversarial perturbations mostly by utilizing the C\&W formulation and its variants. In particular, in the deep reinforcement learning setup the C\&W formulation is,
\begin{equation}
\mathnormal{\min_{s_{\textrm{adv}} \in S} c\cdot J(s_{\textrm{adv}}) + \norm{s_{\textrm{adv}}-s}_2^2}
\label{carlini}
\end{equation}
where $s$ is the unperturbed input, $s_{\textrm{adv}}$ is the adversarially perturbed input, and $J(s)$ is the augmented cost function used to train the network. Note that in the deep reinforcement learning setup the C\&W formulation finds the minimum distance to a decision boundary in which the action taken by the deep reinforcement learning policy is non-optimal. The second method we use to produce adversarial examples is the ENR method \citet{chen18ead},
\begin{equation}
\mathnormal{\min_{s_{\textrm{adv}} \in S} c\cdot  J(s_{\textrm{adv}}) + \lambda_1\norm{s_{\textrm{adv}}-s}_1 + \lambda_2\norm{s_{\textrm{adv}}-s}_2^2}
\label{ead}
\end{equation}
By adding ENR this method produces sparser perturbations compared to the C\&W formulation with similar $\ell_2$-norm.

\subsection{Deep Reinforcement Learning and Adversarial Perspective}

The first work on adversarial examples in deep reinforcement learning appeared in \citet{huang17} and \citet{kos17}. These two concurrent papers demonstrated that FGSM adversarial examples could significantly decrease the performance of deep reinforcement learning policies in standard baselines.
Follow up work by \citet{pattanaik17} introduced an alternative to the standard FGSM by utilizing an objective function which attempts to increase the probability of the worst possible action (i.e. the action $a_w$ minimizing $Q(s,a)$).

The work of \citet{lin17} and \citet{sun20} attempts to exploit the difference in the importance of states across time by only introducing adversarial perturbations at strategically chosen moments. In these papers the perturbations are computed via the C\&W formulation.
Due to the vulnerabilities outlined in the papers mentioned above, there has been another line of work on attempting to train agents that are robust to adversarial perturbations. Initially \citet{mandlekar17} introduced adversarial examples during reinforcement learning training to improve robustness. An alternate approach of \citet{pinto17} involves modelling the agent and the adversary as playing a zero-sum game, and training both
agent and adversary simultaneously.
A similar idea of modelling the agent vs the adversary appears in \citet{glaeve19}, but with an alternative constraint that the adversary must take natural actions in the simulated environment rather than introducing $\ell_p$-norm bounded perturbations to the agent's observations.
Most recently, \citet{huan20} introduced the notion of a State-Adversarial MDP, and used this new definition to design an adversarial training algorithm for deep reinforcement learning with more solid theoretical motivation.
Subsequently, several concerns have been raised regarding the problems introduced by adversarial training \cite{korkmaz21nips, korkmaz21cvpr, korkmazuai}.

\section{High-Sensitivity Directions}
\label{highsen}
The approach in our paper is based on identifying directions of high sensitivity for neural policies. A first attempt to define the notion of high-sensitivity direction might be to say that $\vv$ is a high sensitivity direction for $\pi$ at state $s$ if for some small $\epsilon > 0$
\[
\argmax_a \pi(s + \epsilon \vv,a) \neq \argmax_a \pi(s,a)
\]
but for a random direction $\vr = \frac{\norm{\vv}}{\norm{\vg}}\vg$ with $\vg \sim \gN\left(0,I\right)$
\[
\argmax_a \pi(s + \epsilon \vr,a) = \argmax_a \pi(s,a)
\]
with high probability over the random vector $\vr$.
In other words, $\vv$ is a high sensitivity direction if small perturbations along $\vv$ change the action taken by policy $\pi$ in state $s$, but perturbations of the same magnitude in a random direction do not.
 There is a subtle issue with this definition however.

In reinforcement learning there are no ground truth labels for the correct action for an agent to take in a given state. While the above definition requires that the policy switches which action is taken when the input is shifted a small amount in the $\vv$ direction, there is no \textit{a priori} reason that this shift will be bad for the agent.

Instead, the only objective metric we have of the performance of a policy is the final reward obtained by the agent. Therefore we define high-sensitivity direction as follows:
\begin{definition}
	\label{def:sensitivity}
	Recall that $R = \sum_{t=0}^{T-1}\gamma^tr(s_t,a_t)$ is the cumulative reward.
	A vector $\vv$ is a \emph{high-sensitivity direction} for a policy $\pi$ if there is an $\eps > 0$ such that
	\[
		\E_{a_t \sim \pi(s_t + \eps\vv,\cdot)}[R] \ll \E_{a_t \sim \pi(s_t,\cdot)}[R]
	\]
	but for a random direction $\vr = \frac{\norm{\vv}}{\norm{\vg}}\vg$ with $\vg \sim \gN\left(0,I\right)$
	\[
			\E_{a_t \sim \pi(s_t + \eps\vr,\cdot)}[R] \approx \E_{a_t \sim \pi(s_t,\cdot)}[R].
	\]
\end{definition}
In short, $\vv$ is a high-sensitivity direction if small perturbations along $v$ significantly reduce the expected cumulative rewards of the agent, but the same magnitude perturbations along random directions do not.
This definition ensures not only that small perturbations in the direction $\vv$ cross the decision boundary of the neural policy in many states $s$, but also that the change in policy induced by these perturbations has a semantically meaningful impact on the agent.

To see how high-sensitivity directions arise naturally consider the linear setting where we think of $s$ as a vector of features in $\R^n$, and for each action $a$ we associate a weight vector $\vw_a\in\R^n$. The policy in this setting is given by deterministically taking the action $\argmax_{a \in A}\langle\vw_a,s\rangle$ in state $s$. We assume that the weight vectors are not too correlated with each other: $\langle \vw_a,\vw_{a'} \rangle < \alpha \cdot \min\{\norm{\vw_a}^2,\norm{\vw_{a'}}^2\}$ for some constant $\alpha < 1$. We also assume that the lengths of the weight vectors are not too different: for all $a,a'$ $\norm{\vw_a} \leq \beta \norm{\vw_{a'}}$ for some constant $\beta > 1$.

Suppose that there is a set of states $S_1$ accounting for a significant fraction of the total rewards such that: (1) by taking the optimal action $a^*(s)$ for $s \in S_1$ the agent receives reward 1, and (2) there is one action $b$, such that taking action $b$ gives reward $0$ for a significant fraction of states in $S_1$. We claim that, assuming that there is a constant gap between $\langle\vw_{a^*(s)},s\rangle$ and $\langle\vw_a,s\rangle$ for $a \neq a^*(s)$ then $\vv = \vw_b$ is a high-sensitivity direction.
\begin{proposition}
  \label{propo1}
	Assume that there exist constants $c,d > 0$ such that $c < \langle\vw_{a^*(s)},s\rangle - \langle\vw_a,s\rangle < d$ for all $a \neq a^*(s)$ and $s \in S_1$.
	Then $\vw_b$ is a high-sensitivity direction.
\end{proposition}

\begin{proof}
See Appendix \ref{appa}.
\end{proof}

The empirical results in the rest of the paper confirm that high-sensitivity directions do occur in deep neural policies, and further explore the correlations between these directions across states and across MDPs.

\section{Framework for Investigating High-Sensitivity Directions}
\label{sec:advframework}

In this section we introduce a framework to seek answers for the following questions:

\begin{itemize}
\item \textit{Are high sensitivity directions shared amongst states in the same MDP?}
\item \textit{Is there a correlation between high sensitivity directions across MDPs and across algorithms?}
\item \textit{Do non-robust features exist in the deep reinforcement learning training environments?}
\end{itemize}

It is important to note that the goal of this framework is not to demonstrate the already well-known fact that adversarial perturbations are a problem for deep reinforcement learning. Rather, we are interested in determining whether high-sensitivity directions are correlated across states, across MDPs and across training algorithms. The presence of such correlation would indicate that non-robust features are an intrinsic property of the training environments themselves. Thus, understanding the extent to which non-robust features correlate in these ways can serve as a guide for how we should design algorithms and training environments to improve robustness.

Given a policy $\pi$ and state $s$ we compute a direction $\vv(s,\pi(s,\cdot))$ of minimal margin to the decision boundary $\pi(s,\cdot)$ in state $s$. We fix a bound $\kappa$ such that perturbations of norm $\kappa$ in a random direction $\vr$ with $\norm{\vr}_2=1$ have insignificant impact on the rewards of the agent.
\[
			\E_{a_t \sim \pi(s_t + \kappa\vr,\cdot)}[R] \approx \E_{a_t \sim \pi(s_t,\cdot)}[R].
	\]

Our framework is based on evaluating the sensitivity of a neural policy along the direction $\vv$ in a sequence of increasingly general settings. Each setting is defined by (1) the state $s$ chosen to compute $\vv(s,\pi(s,\cdot))$, and (2) the states $s'$ where a small perturbation along $\vv(s,\pi(s,\cdot))$ is applied when determining the action taken.

\begin{definition}
\label{def1}
\textit{Individual state setting}, $\mathcal{A}^{\textrm{individual}}$, is the setting where in each state $s$ a new direction is computed and a perturbation along that direction is applied. In each state $s_i$ we compute
\begin{equation}
 s_i^* = s_i+ \kappa \cdot \dfrac{\vv(s_i,\pi(s_i,\cdot))}{ \lVert\vv(s_i,\pi(s_i,\cdot))\rVert_2}
\end{equation}
and then take an action determined by $\pi(s_i^*,\cdot)$.
\end{definition}
The individual state setting acts as a baseline to which we can compare the expected rewards of an agent with inputs perturbed along a single direction $\vv$. If the decline in rewards for an agent whose inputs are perturbed along a single direction is close to the decline in rewards for the individual state setting, this can be seen as evidence that the direction $\vv$ satisfies the first condition of Definition \ref{def:sensitivity}.
\begin{definition}
\label{def4}
\textit{Episode independent random state setting}, $\mathcal{A}_{\textrm{e}}^{\textrm{random}}$, is the setting where a random state $s$ is sampled from a random episode $e$, and the perturbation $\vv(s,\pi(s, \cdot))$ is applied to all the states visited in another episode $e'$. Sample $e \sim p_{\mathcal{M}}$ and $s \sim e$.  Given an episode $e' \sim p_{\mathcal{M}}$, in each state $s'_i$ of $e'$ we compute
\begin{equation}
s_i^* = s'_i+ \kappa \cdot  \dfrac{\vv(s,\pi(s,\cdot))}{ \lVert\vv(s,\pi(s,\cdot))\rVert_2}
\end{equation}
and then take an action determined by $\pi(s_i^*,\cdot)$.
\end{definition}
The episode independent random state setting is designed to identify high-sensitivity directions $\vv$. By comparing the return in this setting with the case of a random direction $\vr$ as described in Definition \ref{def:sensitivity} we can decide whether $\vv$ is a high-sensitivity direction.

\begin{definition}
\label{def6}
\textit{Environment independent random state setting}, $\mathcal{A}_{\mathcal{M}}^{\textrm{random}}$, is the setting where a random state $s(\mathcal{M})$ is sampled from a random episode of the MDP $\mathcal{M}$, and the perturbation $\vv(s(\mathcal{M}),\pi(s(\mathcal{M}),\cdot))$ is applied to all the states visited in an episode of a different MDP $\mathcal{M}'$. Sample $e \sim \mathcal{M}$ and $s(\mathcal{M}) \sim e$. Given an episode $e' \sim p_{\mathcal{M}'}$, in each state $s'_i$ we compute

\begin{equation}
s_i^* = s'_i+ \kappa  \cdot \dfrac{\vv(s(\mathcal{M}),\pi(s(\mathcal{M}),\cdot)) }{ \lVert\vv(s(\mathcal{M}),\pi(s(\mathcal{M}),\cdot))) \rVert_2}
\end{equation}

and then take an action determined by $\pi(s_i^*,\cdot)$.
\end{definition}

The environment independent random state setting is designed to test whether high-sensitivity directions are shared across MDPs. As with the episode independent setting, comparing with perturbations in a random direction allows us to conclude whether a high-sensitivity direction for $\mathcal{M}$ is also a high-sensitivity direction for $\mathcal{M}'$.

\begin{definition}
\label{def7}
\textit{Algorithm independent random state setting}, $\mathcal{A}_{\textrm{alg}}^{\textrm{random}}$, is the setting where a random state $s(\mathcal{M})$ is sampled from a random episode of the MDP $\mathcal{M}$, and the perturbation $\vv(s(\mathcal{M}),\pi(s(\mathcal{M}),\cdot))$ is applied to all the states visited in an episode for a policy $\pi'$ trained with a different algorithm.
Sample $e \sim \mathcal{M}$ and $s(\mathcal{M}) \sim e$. Given an episode $e' \sim p_{\mathcal{M}'}$, in each state $s'_i$ we compute

\begin{equation}
s_i^* = s'_i+ \kappa  \cdot \dfrac{\vv(s(\mathcal{M}),\pi(s(\mathcal{M}),\cdot)) }{ \lVert\vv(s(\mathcal{M}),\pi(s(\mathcal{M}),\cdot))) \rVert_2}
\end{equation}

and then take an action determined by $\pi'(s_i^*,\cdot)$.
\end{definition}
Note that in the above definition we allow $\mathcal{M} = \mathcal{M'}$, which corresponds to transferring perturbations between training algorithms in the same MDP. We will refer to the setting where $\mathcal{M} \neq \mathcal{M'}$ (i.e. transferring between algorithms and between MDPs at the same time) as algorithm and environment independent, and denote this setting by  $\mathcal{A}_{\textrm{alg}+\mathcal{M}}^{\textrm{random}}$. Algorithm \ref{alg:highsen} gives the implementation for Definition \ref{def7}.

\begin{algorithm}
\caption{High-sensitivity directions with $\mathcal{A}_{\textrm{alg}}^{\textrm{random}}$}\label{alg:highsen}
\textbf{Input:} Episode $e$ of MDP $\mathcal{M}$, state $s(\mathcal{M}) \sim e$, policy $\pi(s, \cdot)$, policy $\pi'(s,\cdot)$, episode $e'$ of MDP $\mathcal{M'}$, perturbation bound $\kappa$

\begin{algorithmic}
\FOR{$s'_i$ in $e'$ }
\STATE{$s_i^* \gets s'_i+ \kappa \cdot \dfrac{\vv(s(\mathcal{M}),\pi(s(\mathcal{M}),\cdot))}{ \lVert\vv(s(\mathcal{M}),\pi(s(\mathcal{M}),\cdot))\rVert_2}$}
\STATE{$a^*(s_i^*) = \argmax_a \pi'(s_i^*,a)$}
\ENDFOR
\RETURN $\E_{a^*(s_i^*)}[R]$
\end{algorithmic}
\end{algorithm}

\begin{table*}[t]
\caption{Impacts of C\&W and ENR formulation for the proposed framework for investigating high sensitivity directions cross-states and cross-MDPs.}
\label{car}
\centering
\scalebox{0.9}{
\begin{tabular}{lccccccr}
\toprule
Settings [Adversarial Technique]    & BankHeist
                           			    & RoadRunner
                        						 & JamesBond
                          				 & CrazyClimber
                          				 &  TimePilot
                          				 &  Pong \\
\midrule
$\mathcal{A}^{\textrm{individual}}$  [ENR] &  0.646$\pm$0.018
      																			& 0.821$\pm$ 0.046
      																			& 0.098$\pm$0.047
      																			& 0.750$\pm$0.030
      																			& 0.815$\pm$0.049
      																			& 0.995$\pm$0.003\\
$\mathcal{A}^{\textrm{individual}}$  [C\&W] &  0.694$\pm$0.045
      																			& 0.876$\pm$0.017
      																			& 0.038$\pm$0.050
      																			& 0.646$\pm$0.056
      																			& 0.334$\pm$0.107
      																			& 1.0$\pm$0.000 \\
\midrule
$\mathcal{A}_{\textrm{e}}^{\textrm{random}}$[ENR]  &  0.764$\pm$0.022
                                    																			& 0.961$\pm$0.008
                                    																			& 0.612$\pm$0.040
                                    																			&  0.980$\pm$ 0.002
                                    																			&0.517$\pm$0.085
                                    																			& 1.0$\pm$0.000  \\
$\mathcal{A}_{\textrm{e}}^{\textrm{random}}$[C\&W]     &  0.118$\pm$0.041
                            																			& 0.649$\pm$0.041
                            																			& 0.019$\pm$0.058
                            																			&  0.956$\pm$0.003
                            																			& 0.228$\pm$0.097
                            																			& 1.0$\pm$0.000  \\
\midrule
$\mathcal{A}_{\mathcal{M}}^{\textrm{random}}$[ENR] &  0.496$\pm$0.055
              																			& 0.816$\pm$0.034
              																			& 0.910$\pm$0.051
              																			& 0.993$\pm$0.002
              																			& 0.312$\pm$0.118
              																			& 0.946$\pm$0.017  \\
$\mathcal{A}_{\mathcal{M}}^{\textrm{random}}$[C\&W] &  0.022$\pm$0.0349
              																			& 0.919$\pm$0.018
              																			& 0.304$\pm$0.038
              																			& 0.036$\pm$0.028
              																			& 0.017$\pm$0.073
              																			&  0.966$\pm$0.008\\
\midrule
Gaussian    &  0.078$\pm$0.037  & 0.027$\pm$0.025 &  0.038$\pm$0.063 &  0.054$\pm$0.025   &  0.031$\pm$0.063 &  0.045$\pm$0.018    \\
\bottomrule
\end{tabular}
}
\end{table*}

\section{Experiments}
\label{experiments}

The Arcade Learning Environment (ALE) is used as a standard baseline to compare and evaluate new deep reinforcement learning algorithms as they are developed \cite{hasselt15, mnih16, wang16, fedus20, rowland19, mn15, hessel18, kaptur19, dabney18, dabney21, xu20, simon20}. As a result, any systematic issues with these environments are of critical importance. Any bias within the environment that favors some algorithms over others risks influencing the direction of research in deep reinforcement learning for the next several years.
This influence could take the form of diverting research effort away from promising algorithms, or giving a false sense of security that certain algorithms will perform well under different conditions. Thus, for these reasons it is essential to investigate the existence of non-robust features and the correlations between high-sensitivity directions within the ALE.

In our experiments agents are trained with Double Deep Q-Network (DDQN) proposed by \citet{wang16} with prioritized experience replay \citet{schaul15} in the ALE introduced by \citet{bell13} with the OpenAI baselines version \citet{openai}. The state-of-the-art adversarially trained deep reinforcement learning polices utilize the State-Adversarial DDQN (SA-DDQN) algorithm proposed by \citet{huan20} with prioritized experience replay \citet{schaul15}. We average over 10 episodes in our experiments. We report the results with standard error of the mean throughout the paper.
The impact on an agent is defined by normalizing the performance drop as follows
\begin{equation}
\textrm{Impact} = \dfrac{\textrm{Score}_{\textrm{max}} - \textrm{Score}_{\textrm{set}}}{\textrm{Score}_{\textrm{max}} - \textrm{Score}_{\textrm{min}}}.
\end{equation}
Here $\textrm{Score}_{\textrm{max}}$ is the score of the baseline trained agent following the learned policy in a clean run of the agent in the environment, $\textrm{Score}_{\textrm{set}}$ is the score of the agent with settings introduced in Section \ref{sec:advframework}, and $\textrm{Score}_{\textrm{min}}$ is the score the agent receives when choosing the worst possible action in each state. All scores are recorded at the end of an episode.

\subsection{Investigating High-Sensitivity Directions}

Table \ref{car} shows impact values in games from the Atari Baselines for the different settings from our framework utilizing the C\&W formulation, ENR formulation and Gaussian noise respectively to compute the directions $\vv$. In all the experiments we set the $\ell_2$-norm bound $\kappa$ in our framework to a level so that Gaussian noise with $\ell_2$-norm $\kappa$ has insignificant impact. In Table \ref{car} we show the Gaussian noise impacts on the environments of interest with the same $\ell_2$-norm bound $\kappa$ used in the ENR formulation and C\&W formulations. Therefore, high impact for the $\mathcal{A}_{\textrm{e}}^{\textrm{random}}$ and $\mathcal{A}_{\mathcal{M}}^{\textrm{random}}$ setting in these experiments indicates that we have identified a high-sensitivity direction.
The results indicate that it is generally true that a direction of small margin corresponds to a high-sensitivity direction. However, the ENR formulation is more consistent in identifying high-sensitivity directions.

It is extremely surprising to notice that the impact of $\mathcal{A}^{\textrm{individual}}$ in JamesBond is distinctly lower than $\mathcal{A}_{\textrm{e}}^{\textrm{random}}$. The reason for this in JamesBond is that $\mathcal{A}_{\textrm{e}}^{\textrm{random}}$ consistently shifts all actions towards action 12 while $\mathcal{A}^{\textrm{individual}}$ causes the agent to choose different actions in every state. See section \ref{empiricalaction} for more details. We observe that this consistent shift towards one particular action results in a larger impact on the agent's performance in certain environments. In JamesBond, there are obstacles that the agent must jump over in order to avoid death, and consistently taking action 12 prevents the agent from jumping far enough. In CrazyClimber, the consistent shift towards one action results in the agent getting stuck in one state where choosing any other action would likely free it.\footnote{See Appendix \ref{appb} for more detailed information on this issue, and the visualizations of the state representations of the deep reinforcement learning policies for these cases.}

We further investigated the state-of-the-art adversarially trained deep reinforcement learning policies as well as architectural differences in variants of DQN with our framework in Section \ref{advtrain} and in Appendix \ref{para}. We have found consistent results on identifying high-sensitivity directions independent from the training algorithms and the architectural differences.

\subsection{Shifts in Actions Taken}
\label{empiricalaction}

In this subsection we investigate more closely exactly how perturbations along high-sensitivity directions affect the actions taken by the agent.
We then argue that the results of this section suggest that high-sensitivity directions correspond to meaningful non-robust features used by the agent to make decisions. Note that in the reinforcement learning setting it is somewhat subtle to argue about the relationship between perturbations, non-robust features and actions taken. For example, suppose that the direction $\vv$ does indeed correspond to a non-robust feature which the neural policy uses to decide which action to take in several states. The presence of the feature may indicate that different actions should be taken in state $s$ versus state $s'$, even though the feature itself is the same. Thus, even if $\vv$ corresponds to a non-robust feature, this may not be detectable by looking at which actions the agent takes in state $s + \eps\vv$ versus state $s$. However, the converse still holds: if there is a noticeable shift across multiple states towards one particular action $a$ under the perturbation, then this constitutes evidence that the direction $\vv$ corresponds to a non-robust feature that the neural policy uses to decide to take action $a$.

In the OpenAI Atari baseline version of the ALE, the discrete action set is numbered from $0$ to $\abs{\sA}$. For each episode we collected statistics on the probability $P(a)$ of taking action $a$ in the following scenarios:

\begin{itemize}
\label{pdef}
\item $P_{\textrm{base}}(a)$ - the fraction of states in which action $a\in \sA$ is taken in an episode with no perturbation added.
\item $P_{\textrm{shift}}(a)$ - the fraction of states in which action $a\in \sA$ is taken in an episode with the perturbation added.
\item $P_{\textrm{shift}}(a,b)$ - the fraction of states in which action $a\in \sA$ \emph{would have been} taken by the agent if there were no perturbation, but action $b \in \sA$ was taken due to the added perturbation.
\item $P_{\textrm{control}}(a) = \sum_b P_{\textrm{shift}}(a,b)$ - the fraction of states in which action $a\in \sA$ \emph{would have been} taken by the agent if there were no perturbation, in an episode with the perturbation added.
\end{itemize}

\begin{figure}[h!]
\footnotesize
\begin{center}
\stackunder[3pt]{\includegraphics[scale=0.093]{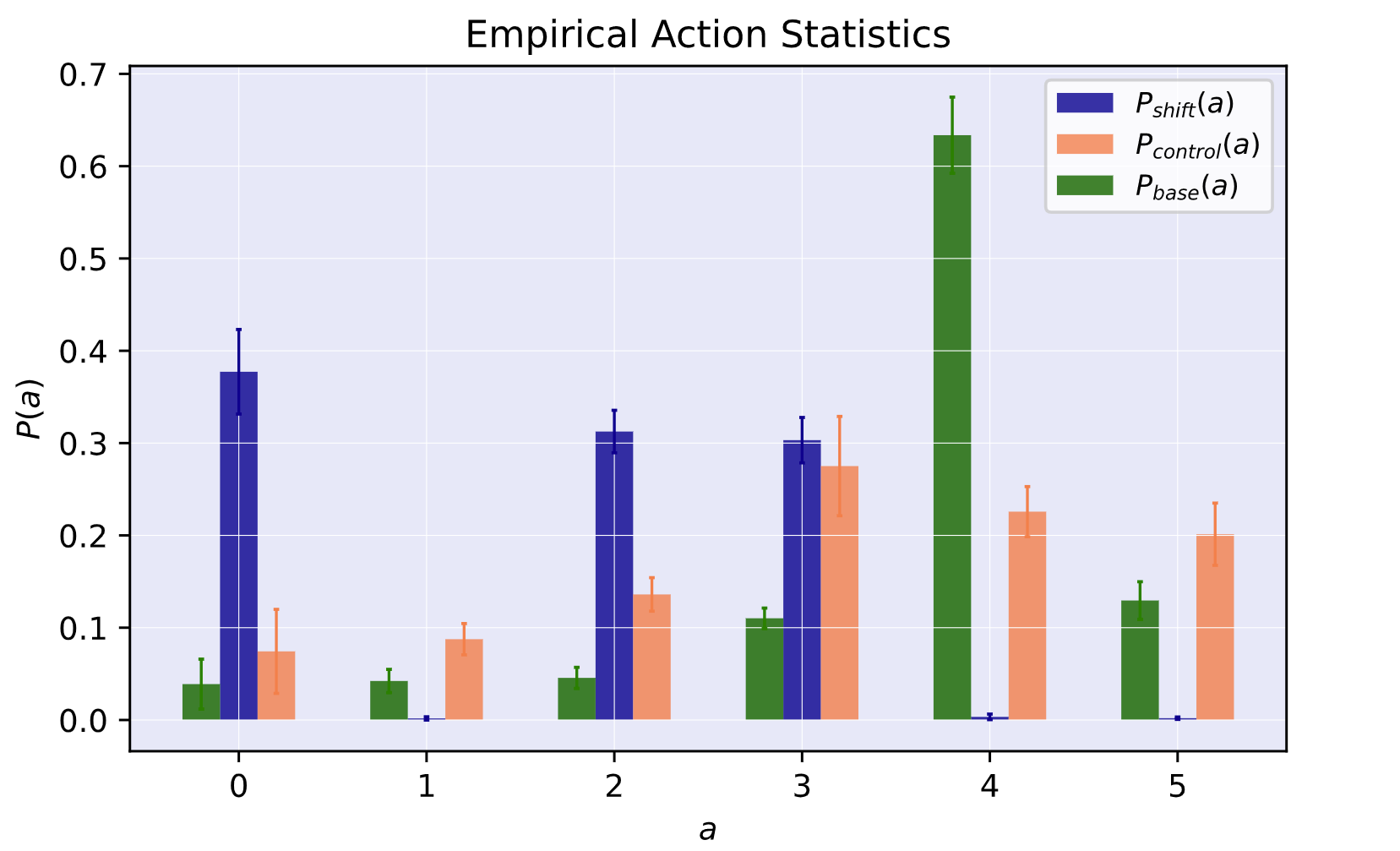}}{Pong}
\stackunder[3pt]{\includegraphics[scale=0.1]{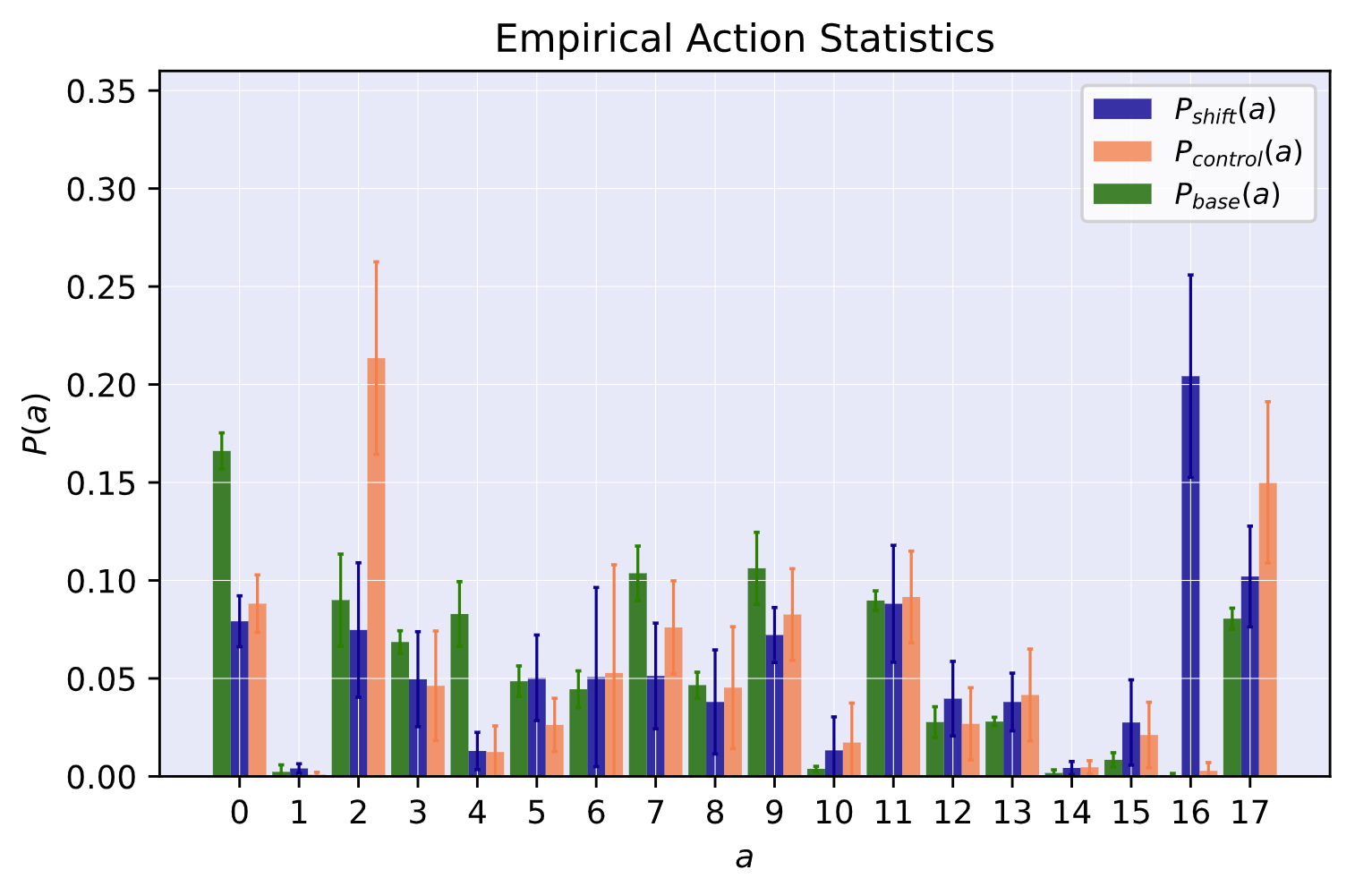}}{BankHeist}
\stackunder[3pt]{\includegraphics[scale=0.095]{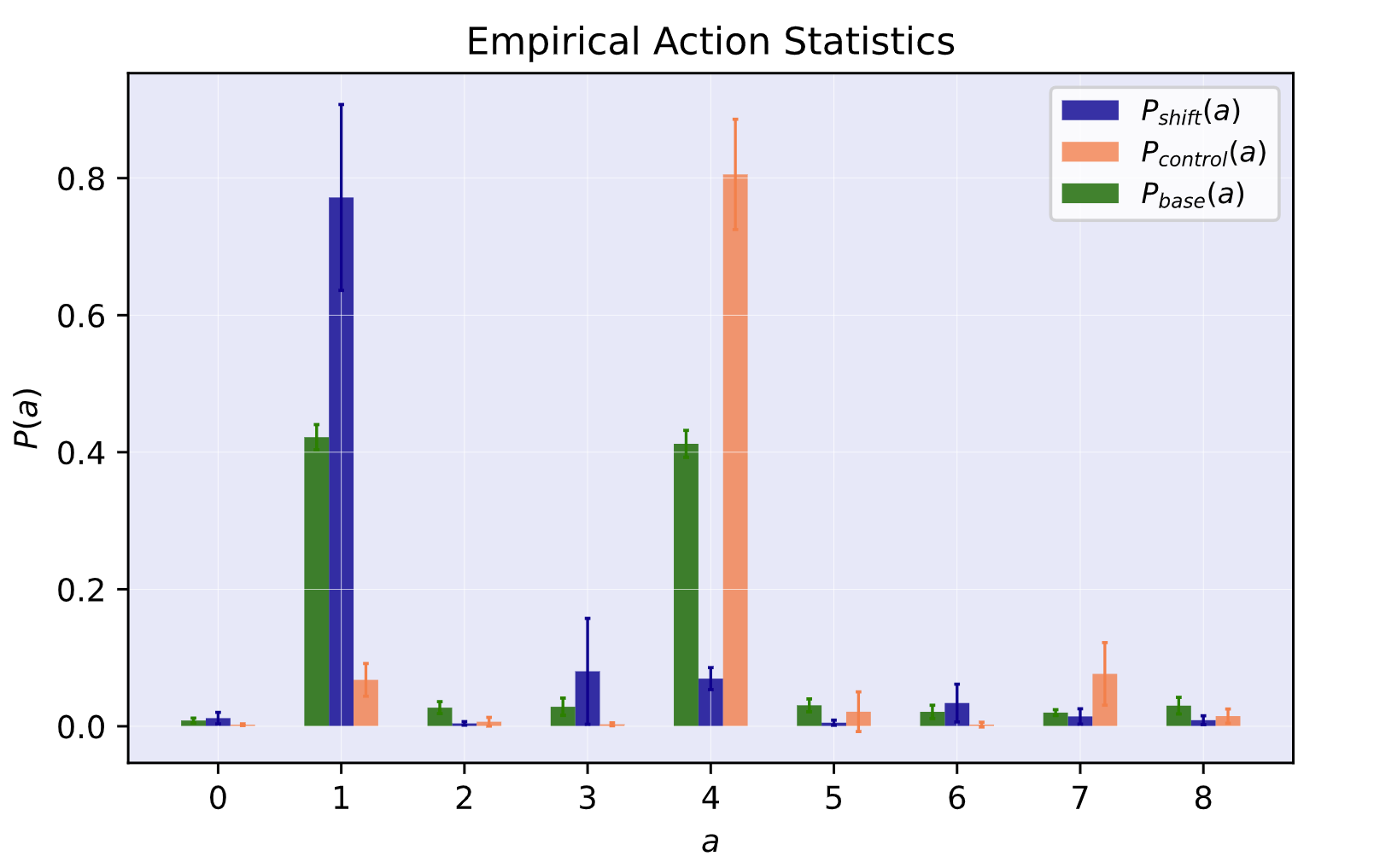}}{CrazyClimber}\\
\stackunder[3pt]{\includegraphics[scale=0.095]{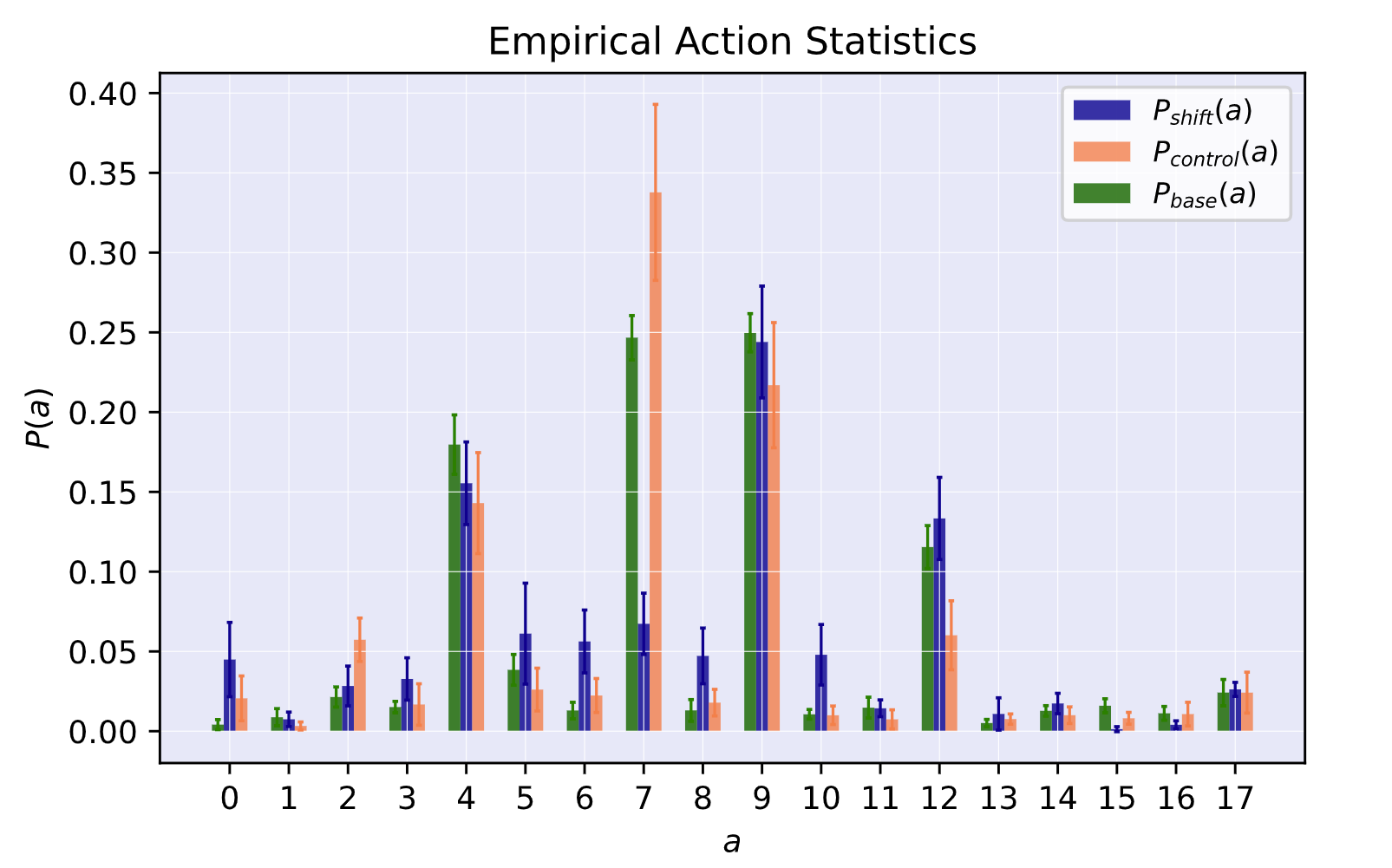}}{RoadRunner}
\stackunder[3pt]{\includegraphics[scale=0.095]{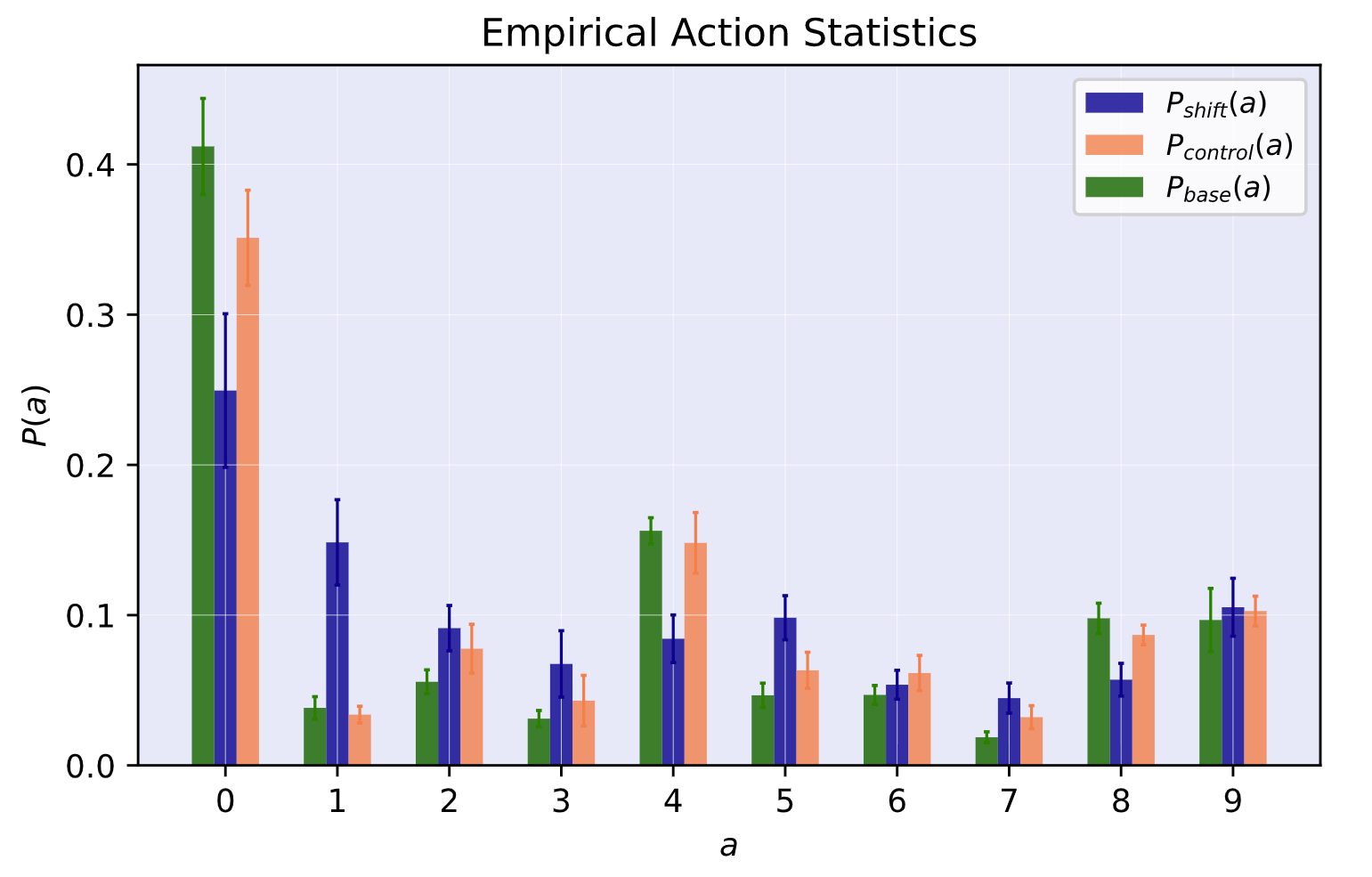}}{TimePilot}
\stackunder[3pt]{\includegraphics[scale=0.098]{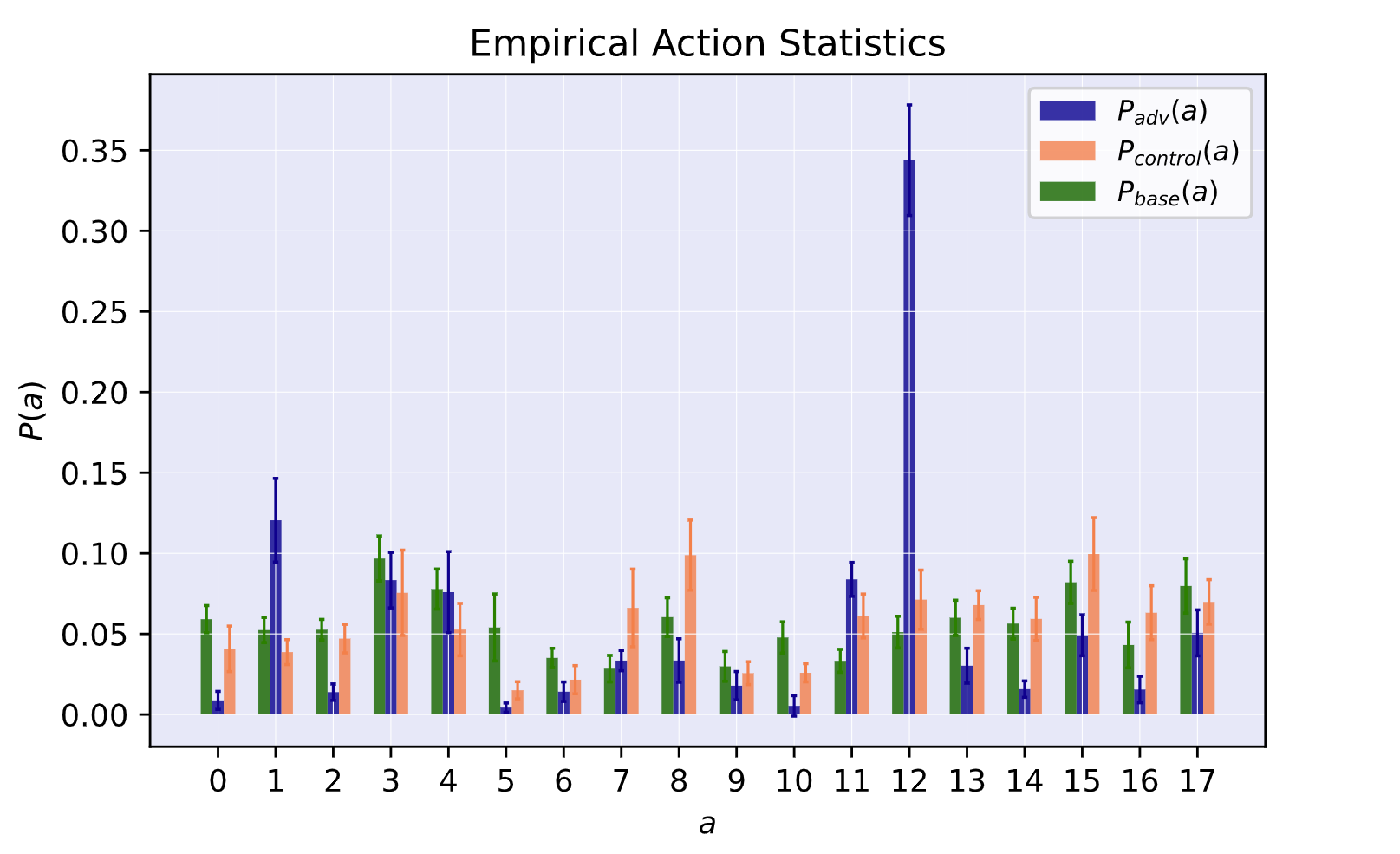}}{JamesBond}
\end{center}
\caption{Action statistics for episode independent random state setting $\mathcal{A}_{\textrm{e}}^{\textrm{random}}$ and environment independent random state setting $\mathcal{A}_{\mathcal{M}}^{\textrm{random}}$ defined in Section \ref{sec:advframework} with ENR formulation for $P_{\textrm{control}}(a)$ , $P_{\textrm{base}}(a)$ and $P_{\textrm{shift}}(a)$.}
\label{actionprob}
\end{figure}

\begin{figure}[t]
\footnotesize
\begin{center}
\stackunder[2pt]{\includegraphics[scale=0.1]{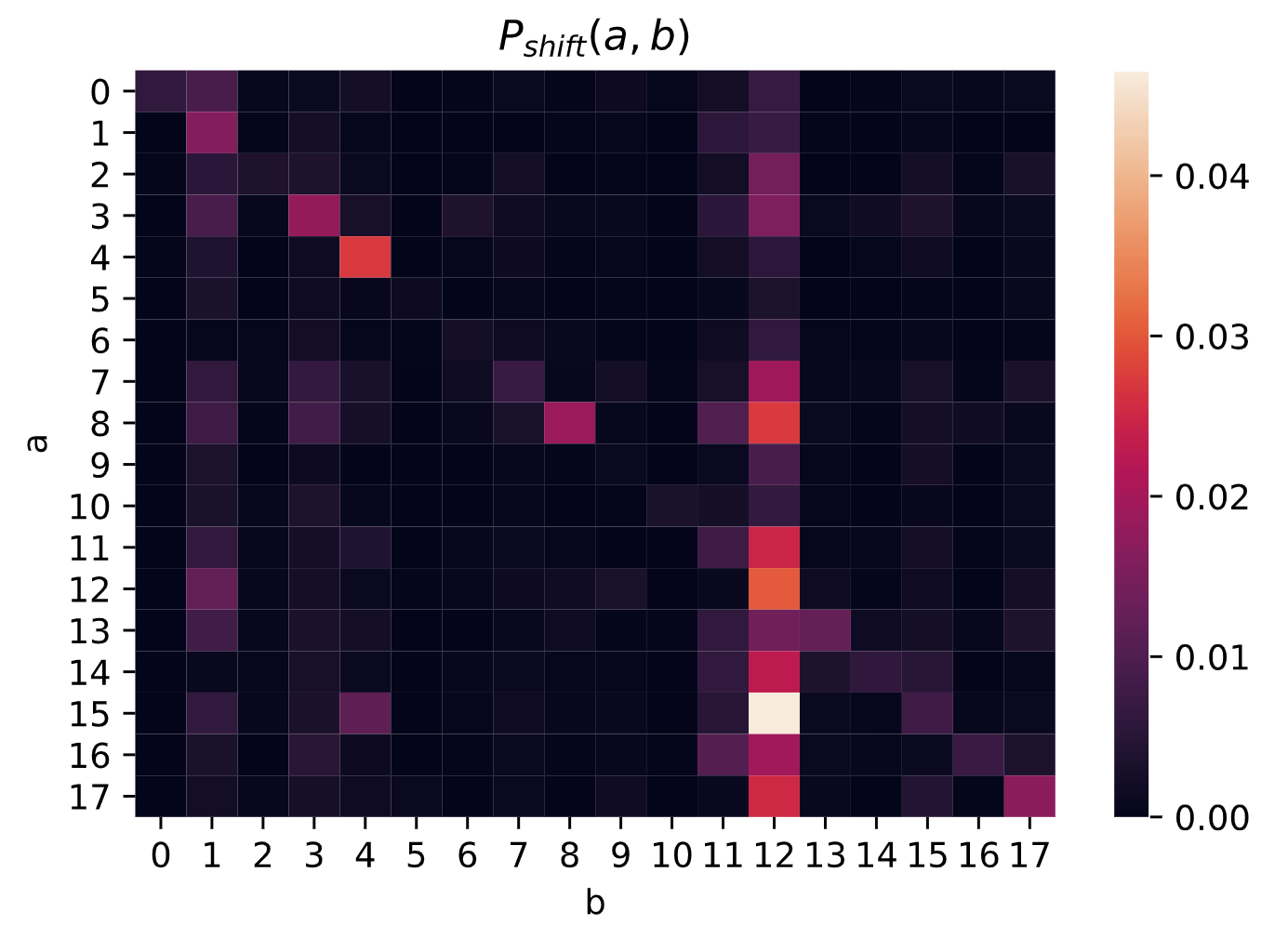}}{JamesBond}
\stackunder[2pt]{\includegraphics[scale=0.1]{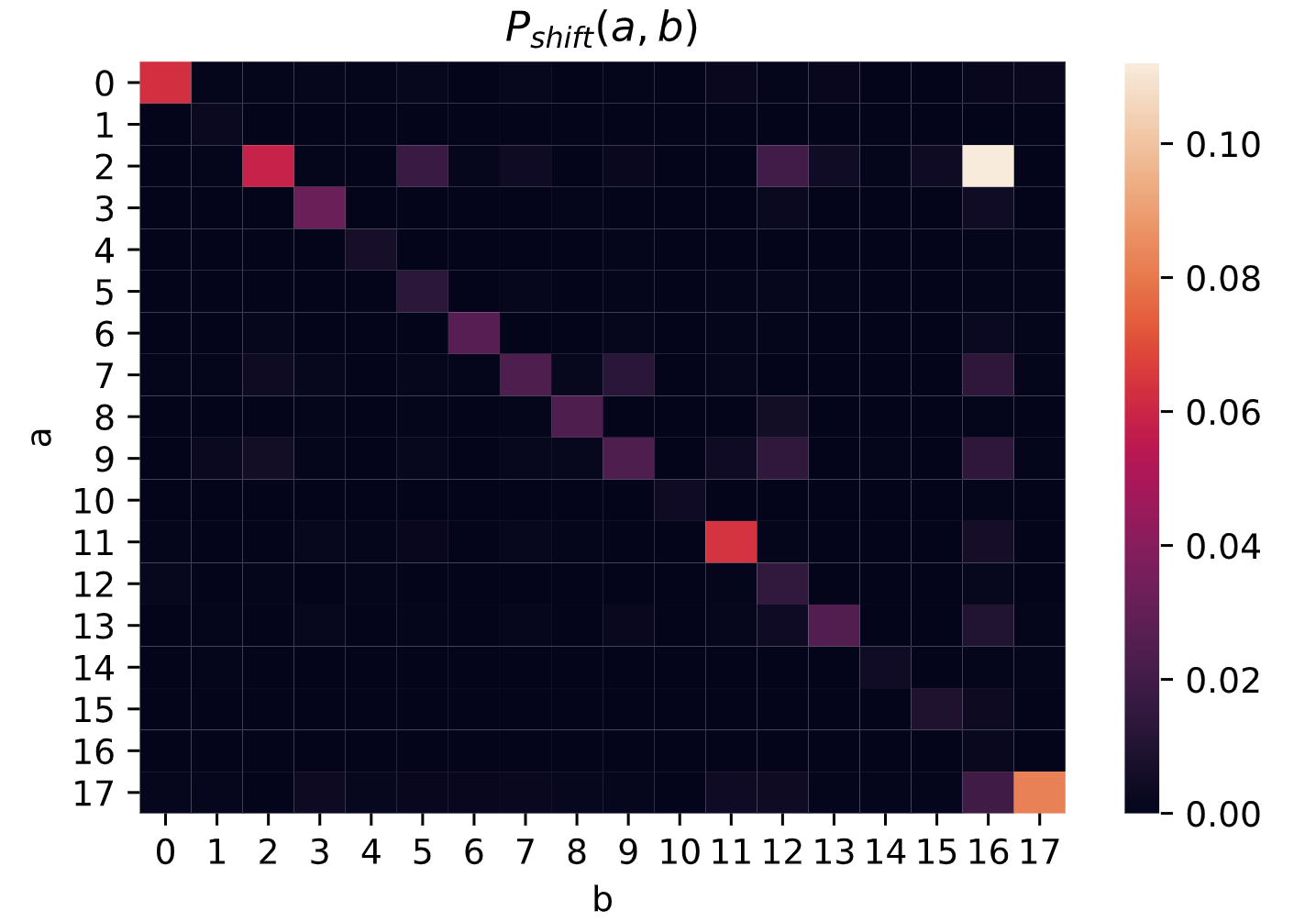}} {BankHeist}
\stackunder[2pt]{\includegraphics[scale=0.1]{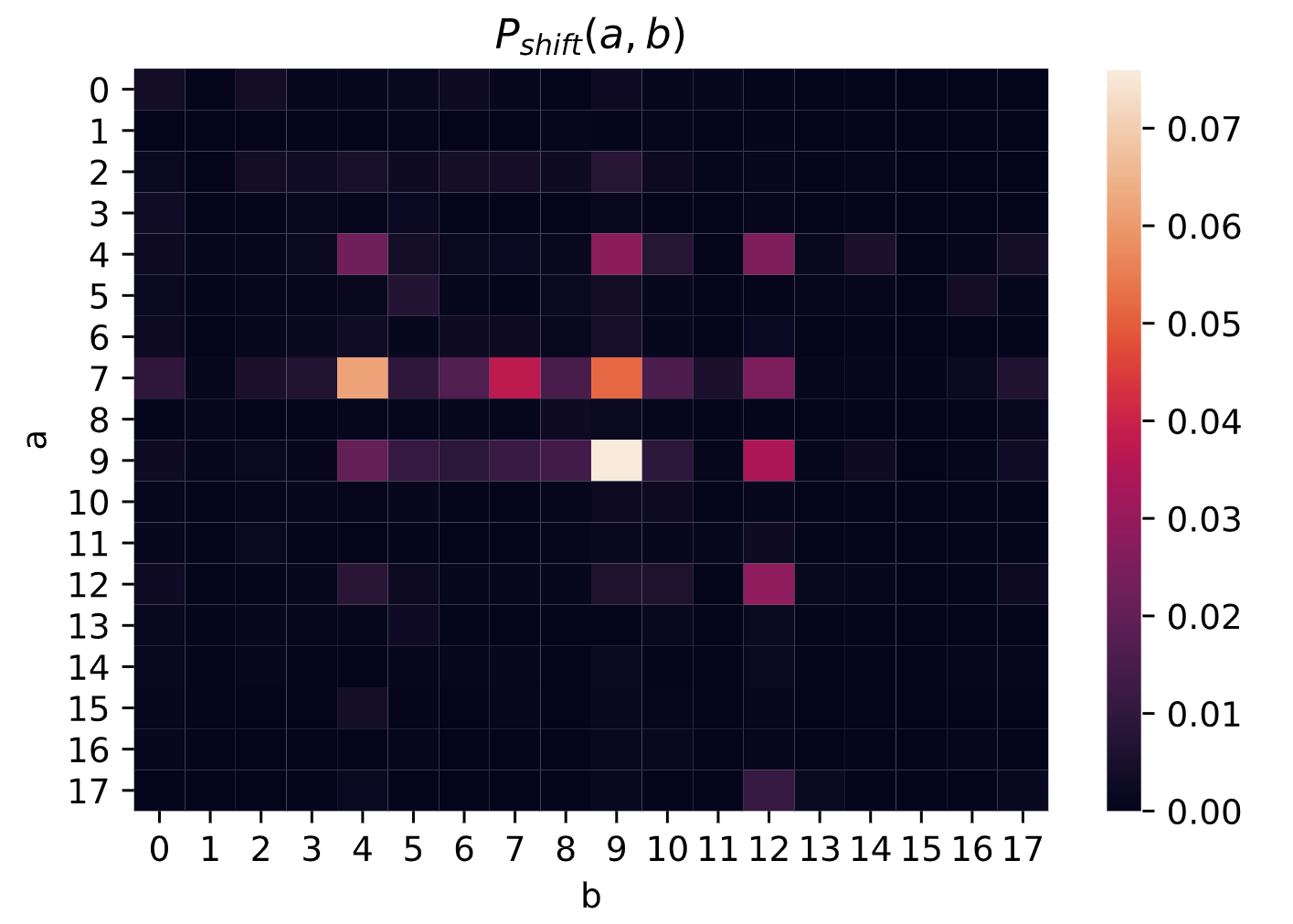}}{RoadRunner}\\
\stackunder[2pt]{\includegraphics[scale=0.11]{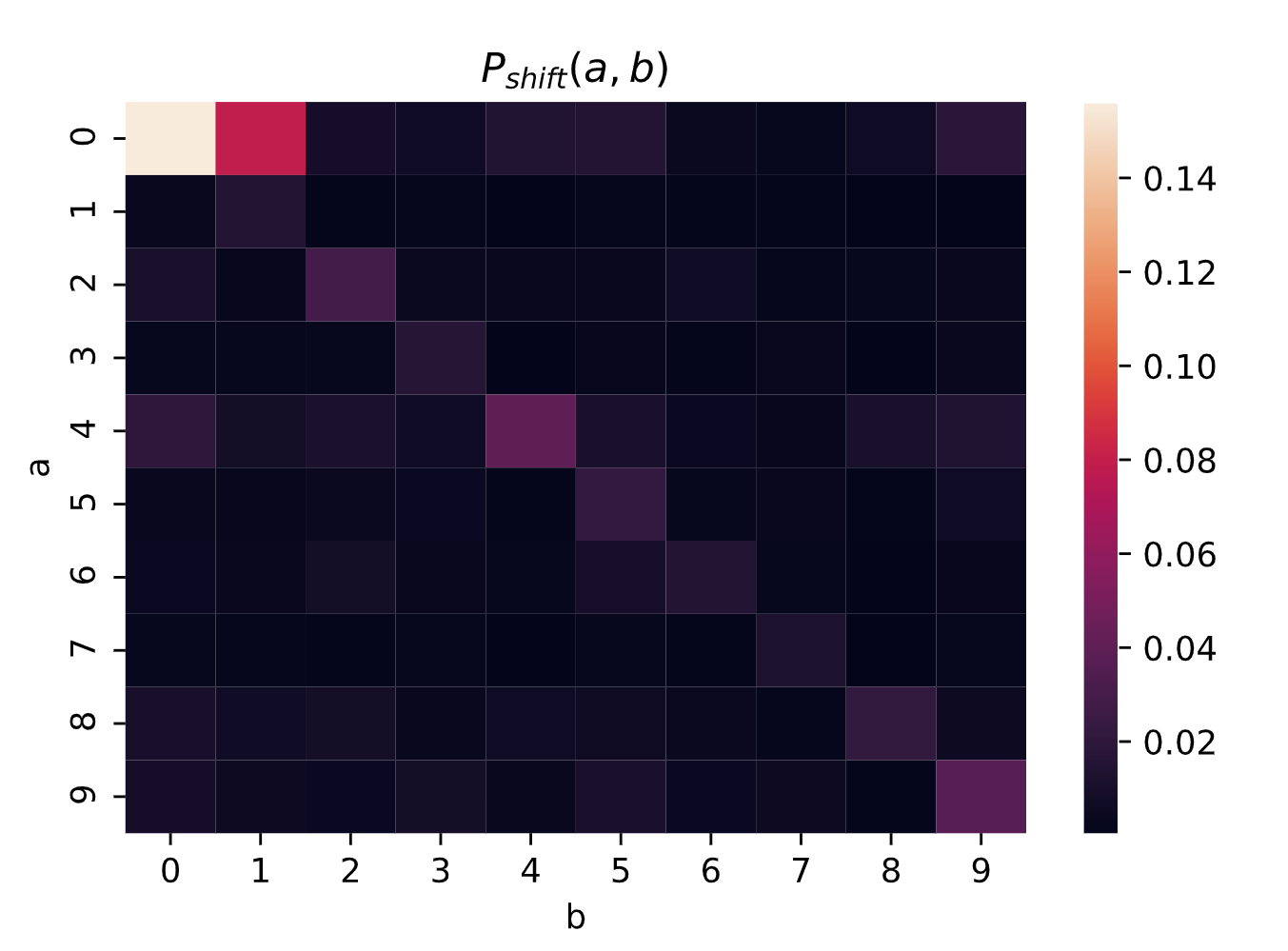}}{TimePilot}
\stackunder[2pt]{\includegraphics[scale=0.1]{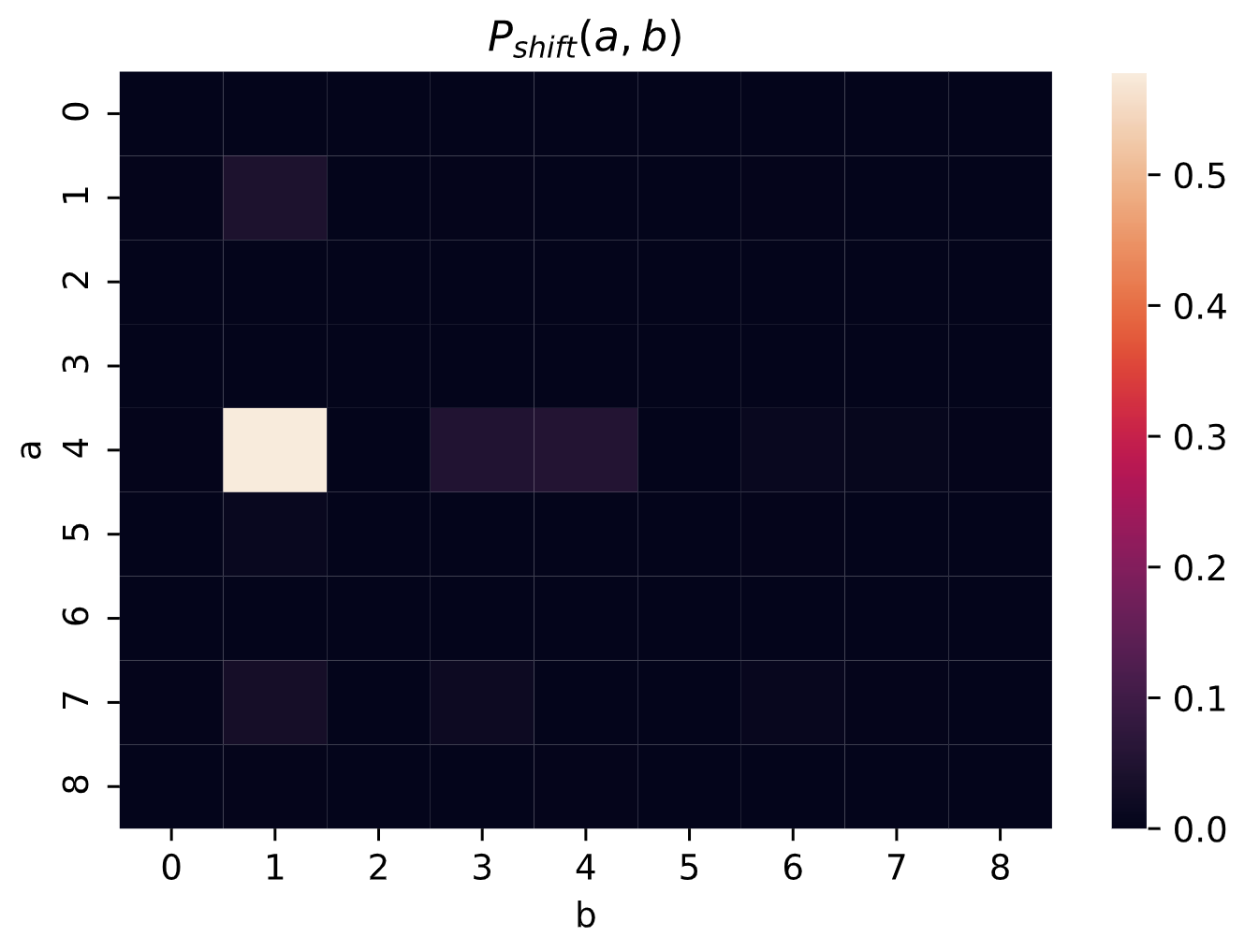}}{CrazyClimber}
\stackunder[2pt]{\includegraphics[scale=0.11]{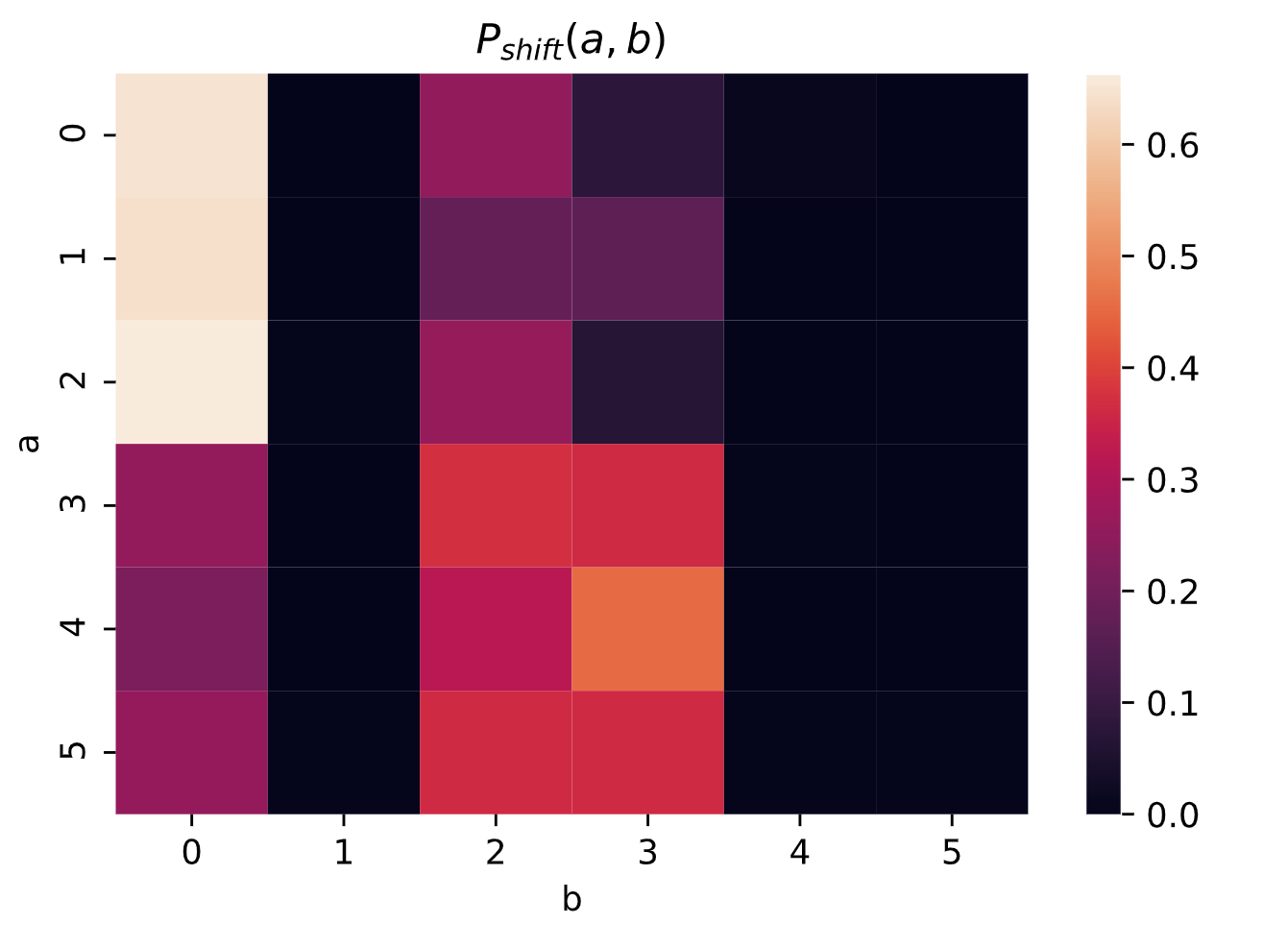}}{Pong}
\end{center}
\caption{Heat map of $P_{\textrm{shift}}(a,b)$ defined in Section \ref{pdef} for episode independent random state setting $\mathcal{A}_{\textrm{e}}^{\textrm{random}}$ and environment independent random state setting $\mathcal{A}_{\mathcal{M}}^{\textrm{random}}$ with ENR}
\label{heatmap}
\end{figure}

In Figure \ref{actionprob} we observe that in Roadrunner, JamesBond, and TimePilot $P_{\textrm{control}}(a)$ and $P_{\textrm{base}}(a)$ are similar. That is, for these environments the distribution on which action \emph{would} be taken without perturbation does not change much. However, significant changes do occur in which action actually is taken due to the perturbation along the high-sensitivity direction. In Pong, BankHeist, and CrazyClimber the presence of the perturbation additionally causes a large change in which action \emph{would} be taken.

In Figure \ref{heatmap} we show the heatmaps corresponding to $P_{\textrm{shift}}(a,b)$. In these heatmaps we show the fraction of states in which the agent would have taken action $a$, but instead took action $b$ due to the perturbation. In some environments there is a dominant action shift towards one particular action $b$ from one particular control action $a$ as in BankHeist, CrazyClimber and TimePilot. The implication for these games is that the high-sensitivity direction $\vv$ corresponds to a non-robust feature that consistently shifts the neural policy's decision from action $a$ to action $b$.

In some games there are one or more clear bright columns where all the control actions are shifted towards the same particular action, for instance in JamesBond.
The presence of a bright column indicates that, for many different input states (where the agent would have taken a variety of different actions), the high-sensitivity direction points across a decision boundary toward one particular action. As argued in the beginning of the section, these results suggest that the high-sensitivity direction found for JamesBond actually corresponds to a meaningful non-robust feature used by the neural policy. Closer examination of the semantics of RoadRunner reveals a similar shift towards a single type of action.

In Roadrunner, the actions 4 and 12 correspond to left and leftfire, both of which move the player directly to the left\footnote{See Appendix \ref{appb} for more details on action names and meanings for each game.}. These actions correspond to two bright columns in the heatmap indicating significant shift towards the actions moving the player directly left from several other actions.
As with the case of JamesBond, this suggests that the high-sensitivity direction $\vv$ found in each of these games corresponds to a non-robust feature that the neural policy uses to decide on movement direction.
To make the results described above more quantitative we report the percentage shift towards a particular action type in Table \ref{percentageshift}. Specifically, we compute the total probability mass $\tau$ on actions which have been shifted from what they would have been, the total probability mass $\rho$ shifted towards the particular action type, and report $\frac{\rho}{\tau}$.

\begin{table}[h!]
\caption{Percentage of total action shift.}
\vskip -0.1in
\label{percentageshift}
\centering
\scalebox{0.9}{
\begin{tabular}{lccccccr}
\toprule
MDP & Action Number & Percentage Shift $\left(\dfrac{\rho}{\tau}\right)$  \\
\midrule
BankHeist & Action 16 & 0.380 \\
RoadRunner &  Actions 4 and 12  & 0.279 \\
JamesBond & Action 12 &  0.345  \\
CrazyClimber &  Action 1 &  0.708 \\
TimePilot &  Action 1  &  0.185 	 \\
\bottomrule
\end{tabular}
}
\end{table}

\subsection{Cross-MDP Correlations in High-sensitivity Directions}
\label{crossmdpcorr}

\begin{figure}[h!]
\footnotesize
\begin{center}
\stackunder[4pt]{\includegraphics[scale=0.16]{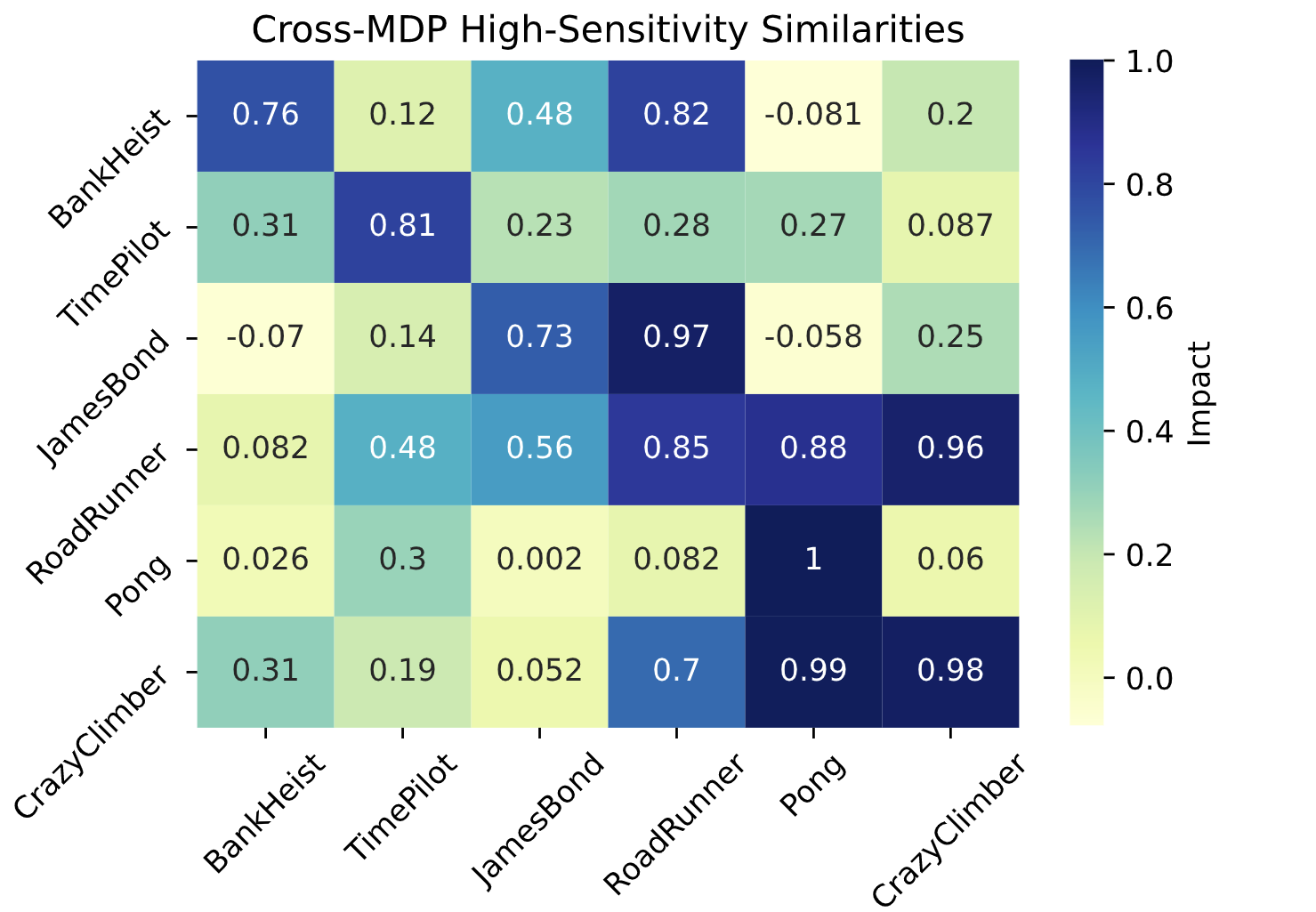}}{}
\stackunder[4pt]{\includegraphics[scale=0.16]{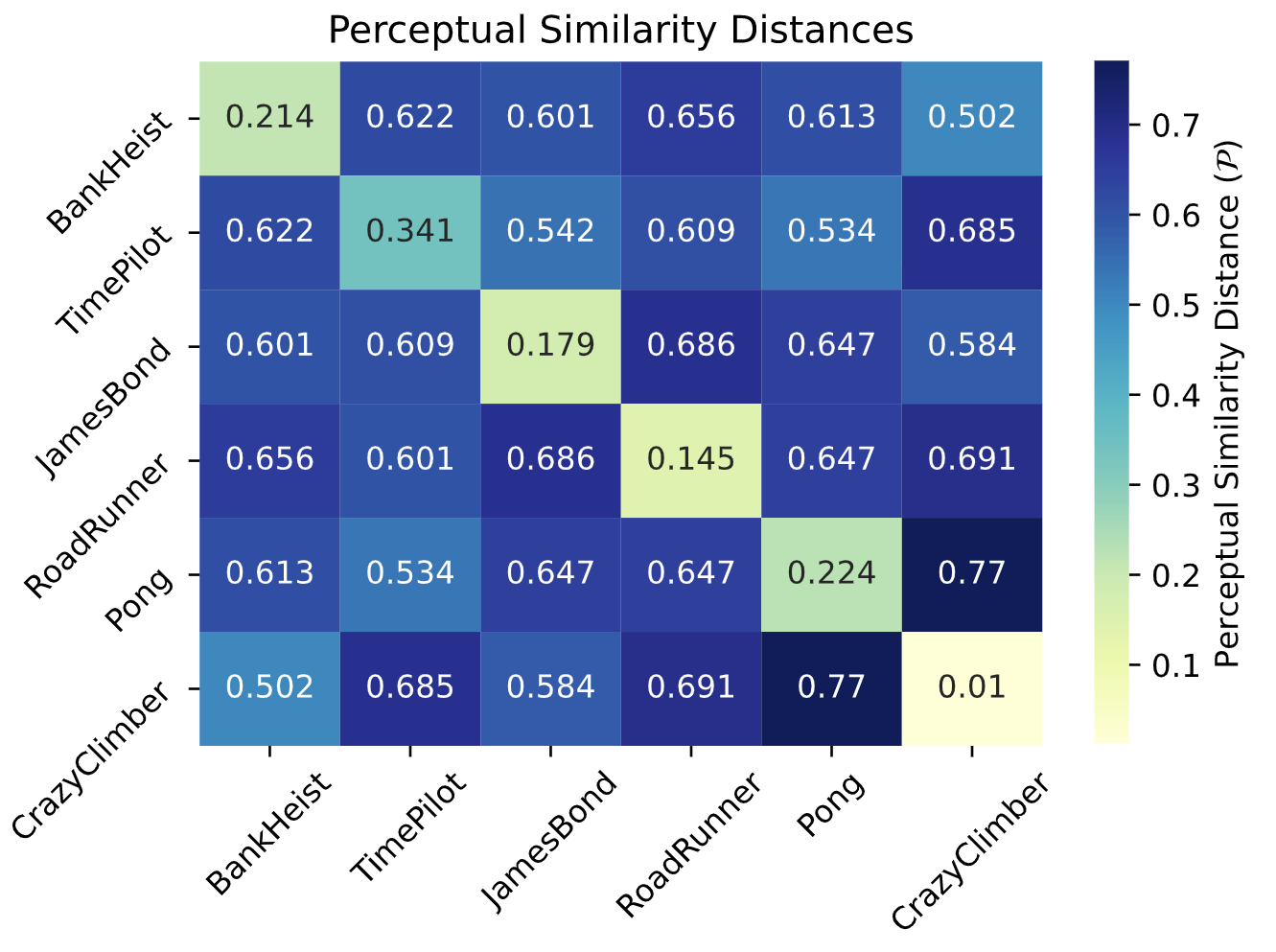}}{}
\end{center}
\caption{Cross-MDP high sensitivity similarities and perceptual similarities for $\mathcal{A}_{\mathcal{M}}^{\textrm{random}}$ with ENR formulation.}
\label{gametrans}
\end{figure}

In this section we investigate more closely the correlation of high-sensitivity directions between MDPs. In these experiments we utilize environment independent random state setting $\mathcal{A}_{\mathcal{M}}^{\textrm{random}}$ where the perturbation is computed from a random state of a random episode of one MDP, and then added to a completely different MDP. In Figure \ref{gametrans} we show the impacts for the $\mathcal{A}_{\mathcal{M}}^{\textrm{random}}$ setting in six different environments. Each row shows the environment where the perturbation is added, and each column shows the environment from which the perturbation is computed. Note that the diagonal of the matrix corresponds to $\mathcal{A}_{e}^{\textrm{random}}$, and thus provides a baseline for the impact value of the high-sensitivity direction in the MDP from which it was computed.
The off diagonal elements represent the degree to which the direction computed in one MDP remains a high-sensitivity direction in another MDP. Perceptual similarity distances are computed via LPIPS \citet{zhang18}.

One can observe intriguing properties of the Atari baselines where certain environments are very likely to share high-sensitivity directions. For example, the high-sensitivity directions computed from RoadRunner are high-sensitivity directions in other MDPs and the high-sensitivity directions computed from other MDPs are high-sensitivity directions for RoadRunner. Additionally, in half of the environments higher or similar impact is achieved when using a direction from a different environment rather than using one computed in the environment itself. Recall that the results of Section \ref{empiricalaction} imply that high-sensitivity directions correspond to non-robust features within a single MDP. Therefore the high level of correlation in high-sensitivity directions across MDPs is an indication that deep reinforcement learning agents are learning representations that have correlated non-robust features across different environments.

\subsection{Shared Non-Robust Features and Adversarial Training}
\label{advtrain}

In this section we investigate the correlations of the high-sensitivity directions between state-of-the-art adversarially trained deep reinforcement learning policies and vanilla trained deep reinforcement learning policies within the same MDP and across MDPs. In particular, Table \ref{adv} demonstrates the performance drop of the policies with settings $\mathcal{A}^{\textrm{Gaussian}}$, $\mathcal{A}_{\textrm{alg}}^{\textrm{random}}$ and $\mathcal{A}_{\textrm{alg}+\mathcal{M}}^{\textrm{random}}$ defined in Section \ref{sec:advframework}.
In more detail, Table \ref{adv} shows that a perturbation computed from a vanilla trained deep reinforcement learning policy trained in the CrazyClimber MDP decreases the performance by 54.6\% when it is introduced to the observation system of the state-of-the-art adversarially trained deep reinforcement learning policy trained in the RoadRunner MDP. Similarly, a perturbation computed from a vanilla trained deep reinforcement learning policy trained in the CrazyClimber MDP decreases the performance by 65.9\% when it is introduced to the observation system of the state-of-the-art adversarially trained deep reinforcement learning policy trained in the Pong MDP. This shows that non-robust features are not only shared across states and across MDPs, but also shared across different training algorithms. The state-of-the-art adversarially trained deep reinforcement learning policies learn similar non-robust features which carry high sensitivity towards certain directions.
It is quite concerning that algorithms specifically focused on solving adversarial vulnerability problems are still learning similar non-robust features as vanilla deep reinforcement learning training algorithms. This fact not only presents serious security concerns for adversarially trained models, but it posits a new research problem on the environments in which we train.

\begin{table}[h!]
\caption{Impacts of $\mathcal{A}^{\textrm{Gaussian}}$, $\mathcal{A}_{\textrm{alg}}^{\textrm{random}}$ and $\mathcal{A}_{\textrm{alg}+\mathcal{M}}^{\textrm{random}}$ where the perturbation is computed from a policy trained with DDQN and introduced to the observation system of the state-of-the-art adversarially trained deep reinforcement learning policy}
\label{adv}
\centering
\scalebox{0.9}{
\begin{tabular}{lccccccr}
\toprule
MDPs           & $\mathcal{A}^{\textrm{Gaussian}}$
      				 & $\mathcal{A}_{\textrm{alg}}^{\textrm{random}}$
      				 & $\mathcal{A}_{\textrm{alg}+\mathcal{M}}^{\textrm{random}}$   \\
\midrule
RoadRunner     &    0.023$\pm$0.058   &    0.397$\pm$0.024      &   0.546$\pm$0.014     \\
Pong           &    0.019$\pm$0.007   &    1.0$\pm$0.000        &   0.659$\pm$0.069     \\
BankHeist      &    0.061$\pm$0.012   &    0.758$\pm$0.042      &   0.241$\pm$0.009     \\
\bottomrule
\end{tabular}
}
\end{table}

\section{Conclusion}

In this paper we focus on several questions: (i) Do neural policies share high sensitivity directions amongst different states? (ii) Are the high sensitivity directions correlated across MDPs and across algorithms? (iii) Do deep reinforcement learning agents learn non-robust features from the deep reinforcement learning training environments? To be able to investigate these questions we introduce a framework containing various settings. Using this framework we show that a direction of small margin to the decision boundary in a single state is often a high-sensitivity direction for the deep neural policy. We then investigate more closely how perturbations along high-sensitivity directions change the actions taken by the agent, and find that in some cases they shift the decisions of the policy towards one particular action. We argue that this suggests that high sensitivity directions correspond to non-robust features used by the policy to make decisions. Furthermore, we show that a high-sensitivity direction for one MDP is likely to be a high-sensitivity direction for another MDP in the Arcade Learning Environment. Moreover, we show that the state-of-the-art adversarially trained deep reinforcement learning policies share the exact same high-sensitivity directions with vanilla trained deep reinforcement learning policies. We systematically show that the non-robust features learnt by deep reinforcement learning policies are decoupled from states, MDPs, training algorithms and the architectural differences. Rather these high-sensitivity directions are a part of the learning environment resulting from learning non-robust features. We believe that this cross-MDP correlation of high-sensitivity directions is important for understanding the decision boundaries of neural policies. Furthermore, the fact that neural policies learn non-robust features that are shared across baseline deep reinforcement learning training environments is crucial for improving and investigating the robustness and generalization of deep reinforcement learning agents.

\bibliography{example_paper}

\newpage

\appendix

\section{High Sensitivity Directions}
\label{appa}

\vspace{0.2cm}

\begin{proposition}
  \label{propo1}
	Assume that there exist constants $c,d > 0$ such that $c < \langle\vw_{a^*(s)},s\rangle - \langle\vw_a,s\rangle < d$ for all $a \neq a^*(s)$ and $s \in S_1$.
	Then $\vw_b$ is a high-sensitivity direction.
\end{proposition}
\begin{proof}
	By assumption for all $a \in A$ and $s \in S_1$ we have
	\[
		  \langle\vw_{a} - \vw_b,s\rangle \leq \langle\vw_{a^*(s)} - \vw_b,s\rangle < d
	\]
	Therefore setting $\eps = \frac{2d}{(1-\alpha)\norm{\vw_b}^2}$ we have
	\begin{align*}
		\langle s + \eps \vw_b, \vw_{a} - \vw_b \rangle &< d + \eps\alpha\norm{\vw_b}^2 - \eps\norm{\vw_b}^2\\
			&= d - (1 - \alpha)\eps\norm{\vw_b}^2 < 0
	\end{align*}
	Thus the when an $\eps$ perturbation is added in the $\vw_b$ direction the agent always takes action $b$ for states $s\in S_1$ and receives zero reward. On the other hand let $\vg \sim \gN\left(0,I\right)$ and set $\vr = \frac{\norm{\vw_b}}{\norm{\vg}}\vg$. For each action $a$, by standard concentration of measure results for uniform random vectors in the unit sphere we have that with probability $1 - \exp(O(-t^2))$
	\[
		\lvert\langle \vr, \vw_a \rangle\rvert < \frac{t\norm{\vw_a}\norm{\vw_b}}{\sqrt{n}}< \frac{t\beta\norm{\vw_a}^2}{\sqrt{n}}
	\]
	Therefore, for each action $a \neq a^*(s)$
	\begin{align*}
		\langle s + \eps \vr, \vw_{a^*(s)} -& \vw_{a} \rangle \\&> c - \eps \frac{t\beta\norm{\vw_{a^*(s)}}^2}{\sqrt{n}} - \eps\frac{t\beta\norm{\vw_a}^2}{\sqrt{n}}\\
		&> c - \frac{4td\beta^3}{(1-\alpha)\sqrt{n}}.
	\end{align*}
	The above lower bound is positive for $t = \frac{c(1-\alpha)\sqrt{n}}{8d\beta^3}$. Since $\alpha$, $\beta$, $c$ and $d$ are constants independent of $n$, we conclude that $\langle s + \eps \vr, \vw_{a^*(s)} - \vw_a \rangle >0$ with probability $1 - \exp(O(-n))$. Therefore, taking a union bound over the set of actions $a\neq a^*(s)$, we have that the agent always takes action $a^*(s)$ in states $s \in S_1$ with high probability, receiving the same rewards as when there is no perturbation.
\end{proof}

The proof of Proposition \ref{propo1} relies primarily on the fact that, when the dimension $n$ is large, for any fixed direction $\vv$, a randomly chosen direction will be nearly orthogonal to $\vv$.
This means that if there is one single direction $\vv$ along which small perturbations can affect the expected cumulative rewards obtained by the policy $\pi$, then random directions are very likely to be orthogonal to $\vv$ and thus very unlikely to have significant impact on rewards.
In the linear setting considered in Proposition \ref{propo1}, one can formally prove this fact. However, it seems plausible that the general phenomenon will carry over to the more complicated setting of deep neural policies, again based on the fact that if there are only a few special high-sensitivity directions $\vv$, then random perturbations are likely to be nearly orthogonal to these directions when the state space is high-dimensional.

\section{Action Mapping and Action Names}
\label{appb}

Table \ref{actionname} shows action names and the corresponding action numbers for RoadRunner. The action names and the action numbers will be used frequently in detail in Section 5.2 in Figure 1 and Figure 2.
\begin{table}[h!]
\caption{Action names and corresponding action numbers for RoadRunner from Arcade Learning Environment \citet{bell13}. The action names and corresponding numbers vary from game to game.}
\label{actionname}
\scalebox{0.9}{
\begin{tabular}{|l|c|c|r|}
\toprule
'NOOP':0       & 'UPRIGHT': 6 &	 'LEFTFIRE':12 \\
'FIRE':1       & 'UPLEFT': 7 &    'DOWNFIRE':13 \\
'UP': 2        & 'DOWNRIGHT': 8 & 'UPRIGHTFIRE':14 \\
'RIGHT ': 3    & 'DOWNLEFT': 9  & 'UPLEFTFIRE':15 \\
'LEFT': 4			 & 'UPFIRE': 10 &  'DOWNRIGHTFIRE':16 \\
'DOWN': 5		   & 'RIGHTFIRE': 11 &  'DOWNLEFTFIRE':17\\
\bottomrule
\end{tabular}
}
\end{table}

\begin{figure}[h!]
\footnotesize
\begin{center}
\stackunder[2pt]{\includegraphics[scale=0.135]{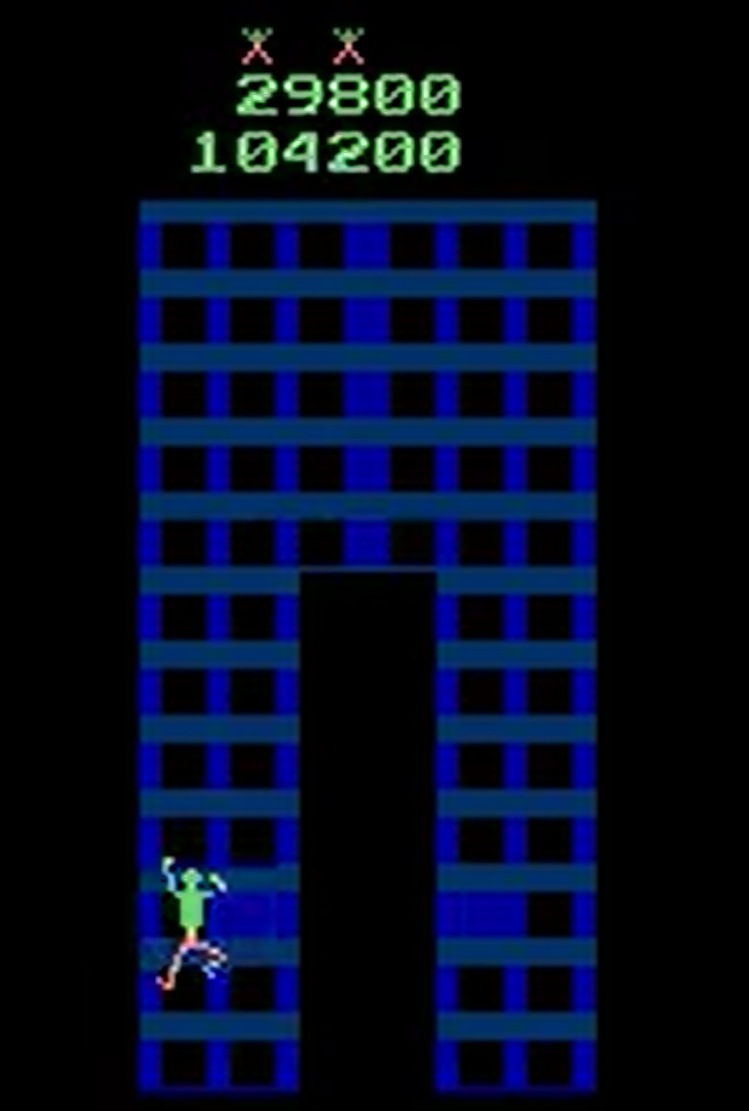}}{}
\stackunder[2pt]{\includegraphics[scale=0.134]{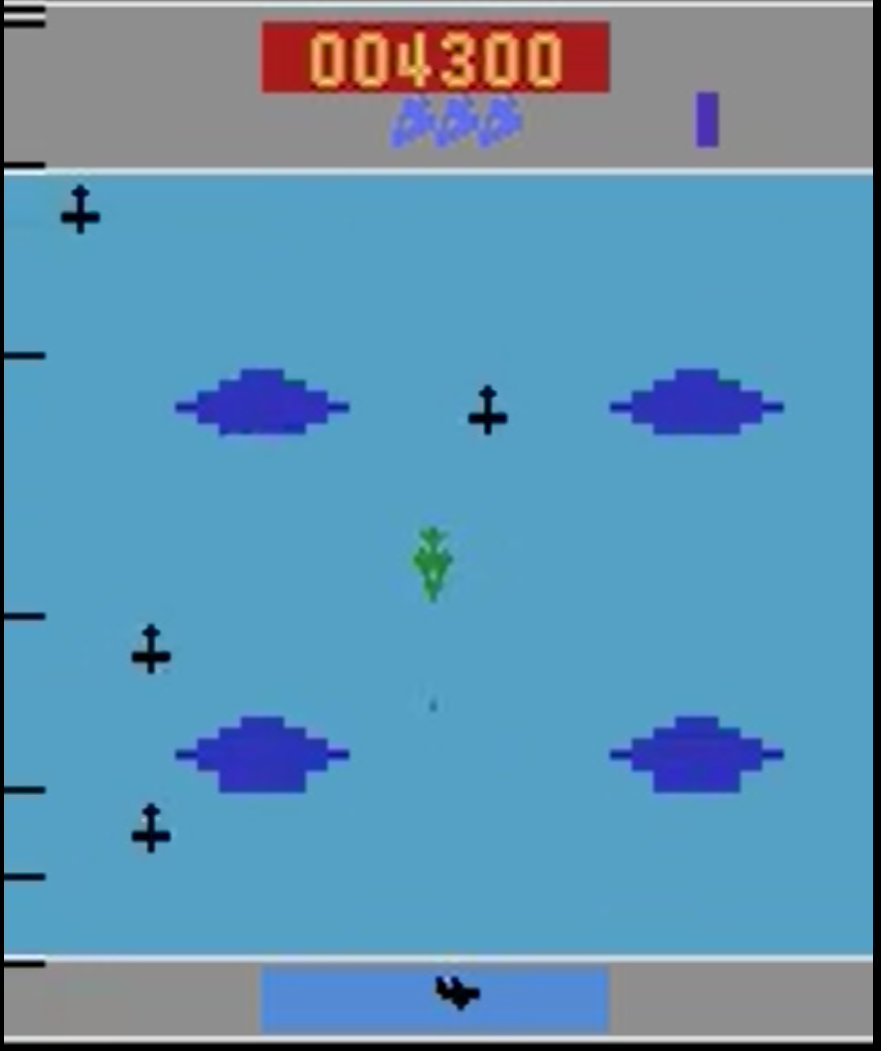}}{}%
\stackunder[2pt]{\includegraphics[scale=0.128]{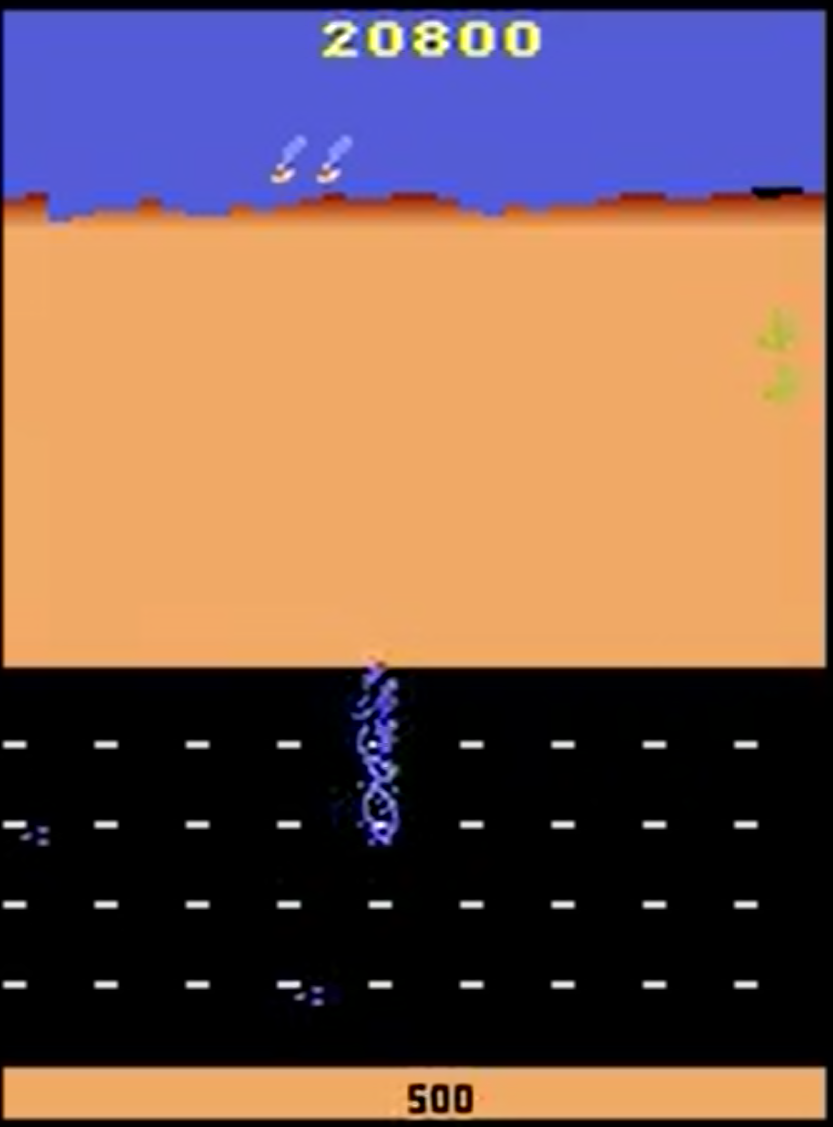}}{}
\stackunder[2pt]{\includegraphics[scale=0.16]{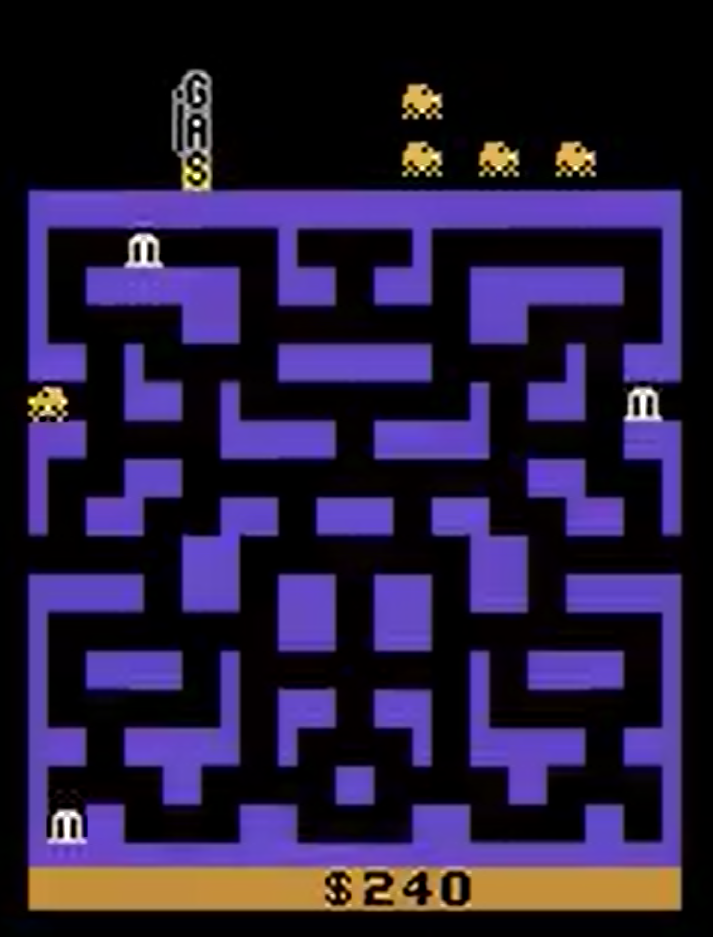}}{}\\
\stackunder[2pt]{\includegraphics[scale=0.135]{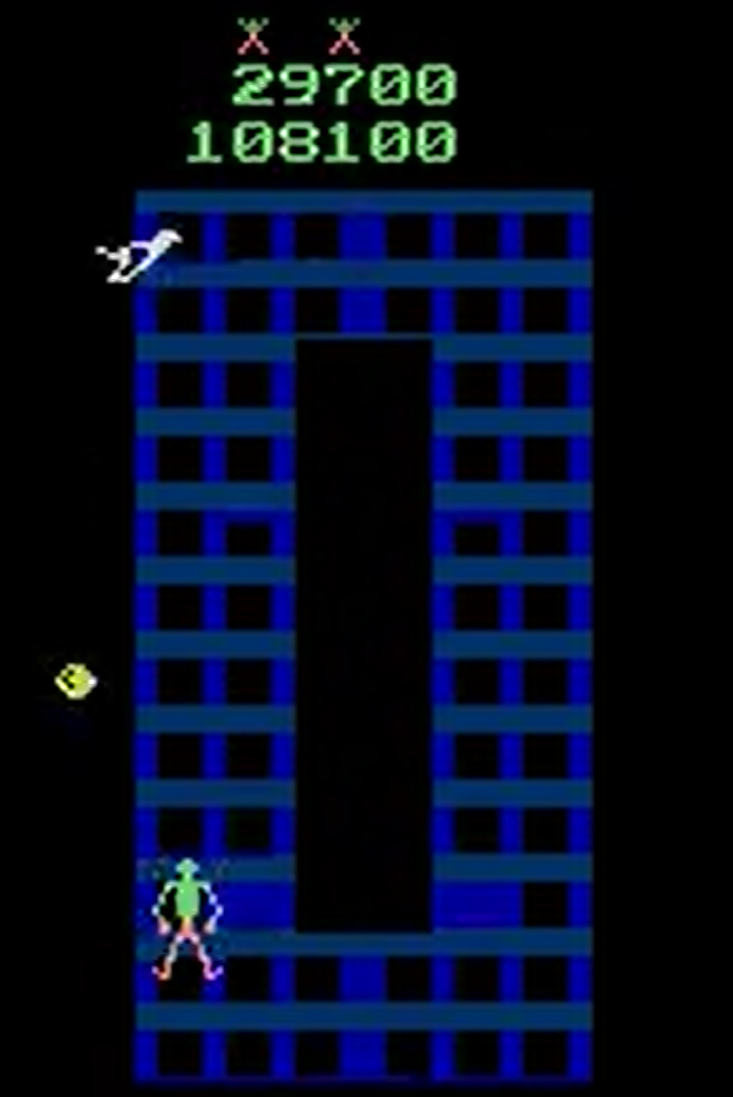}}{}
\stackunder[2pt]{\includegraphics[scale=0.135]{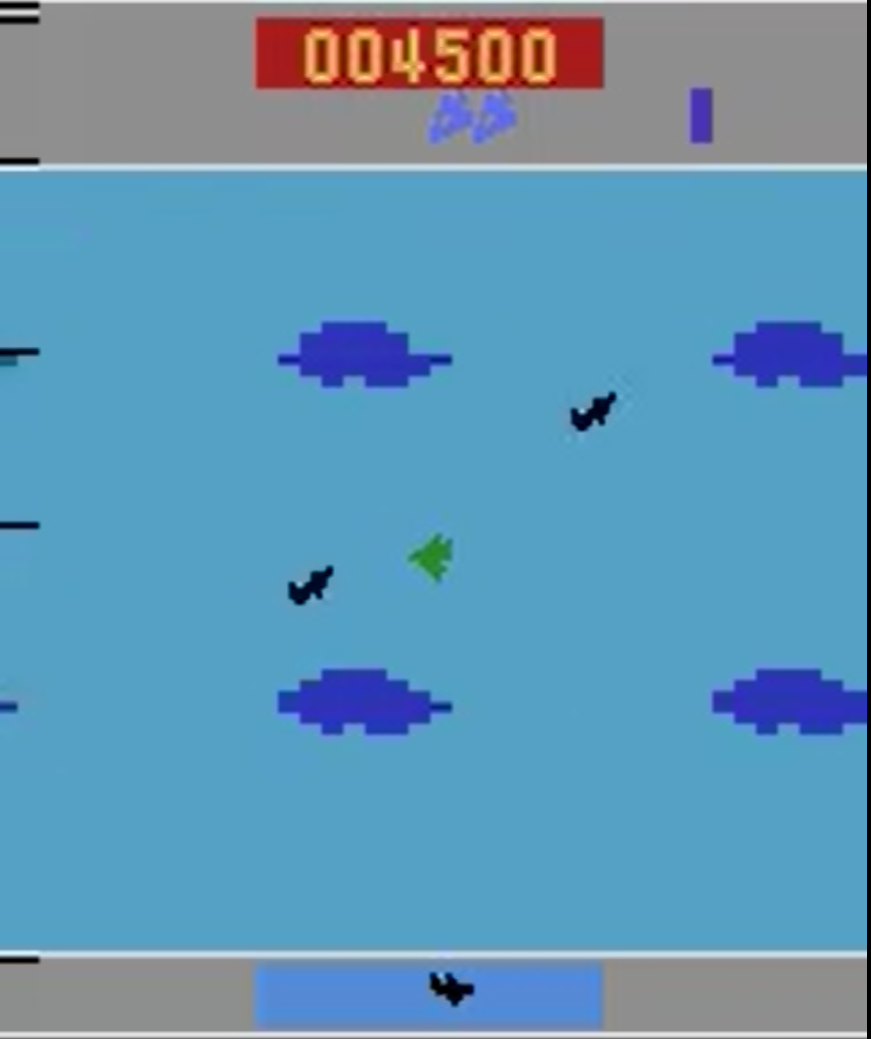}}{}
\stackunder[2pt]{\includegraphics[scale=0.135]{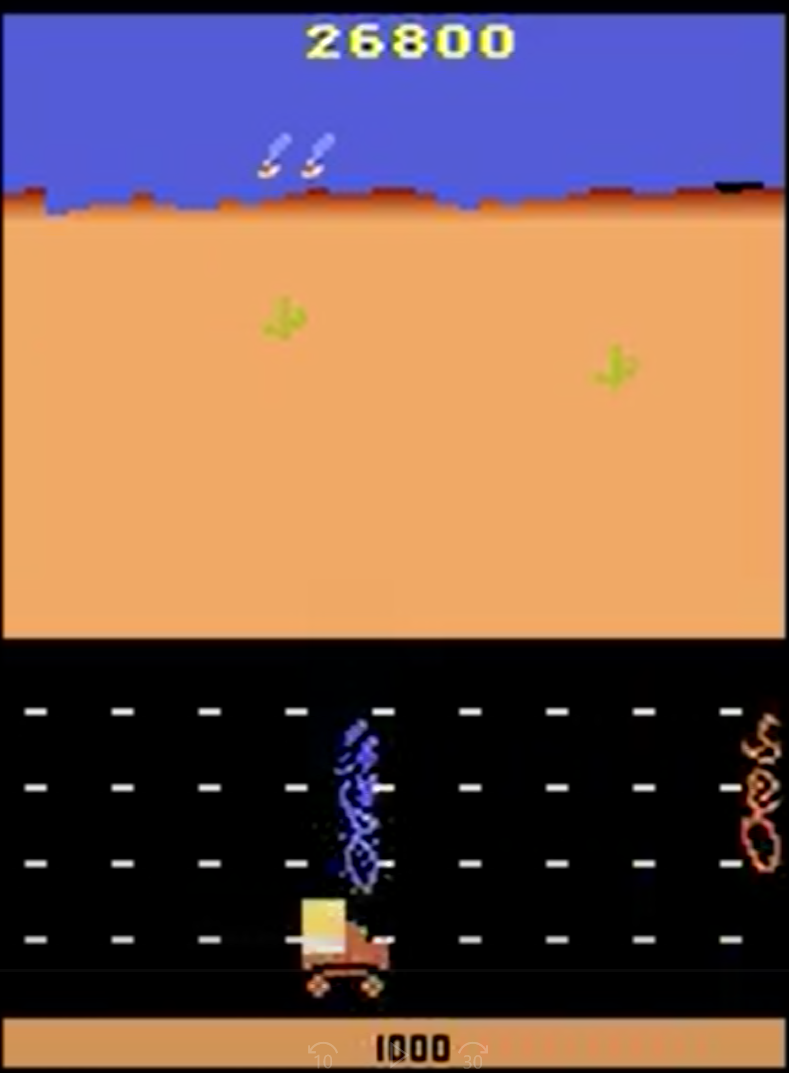}}{}
\stackunder[2pt]{\includegraphics[scale=0.165]{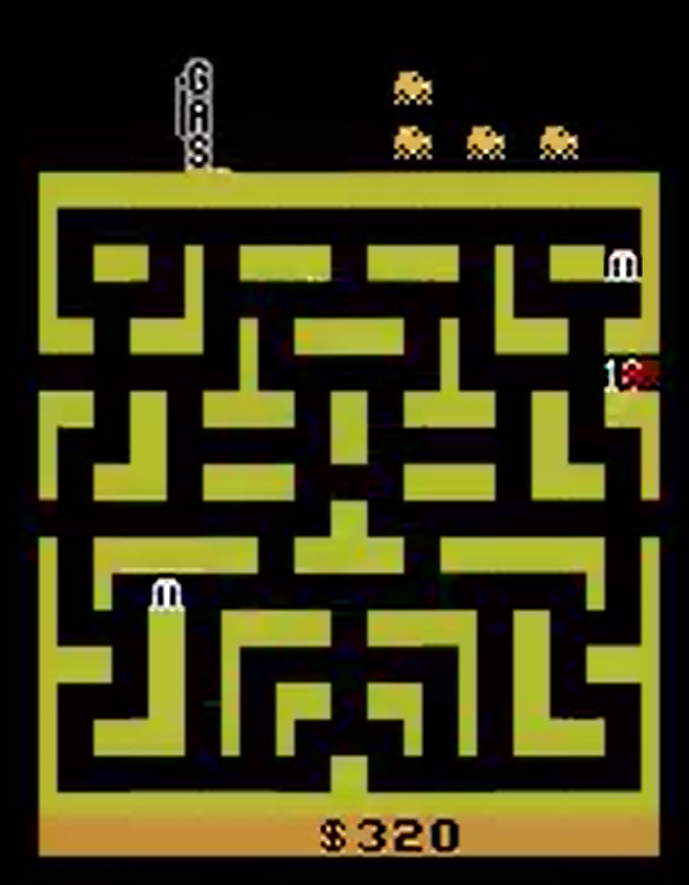}}{}\\
\stackunder[2pt]{\includegraphics[scale=0.14]{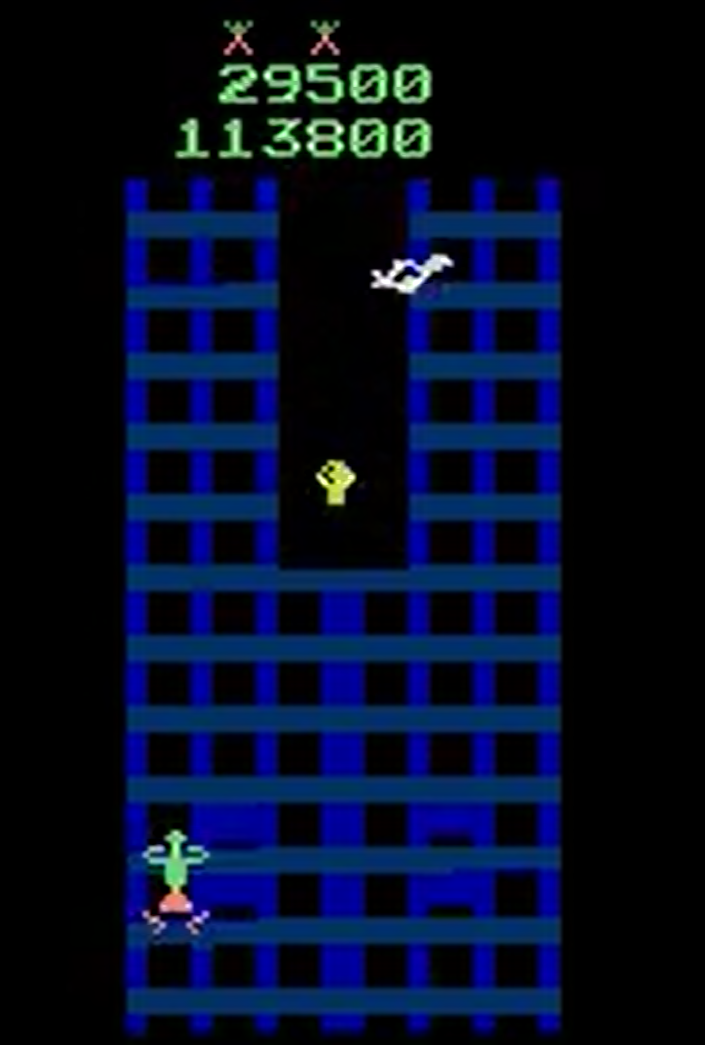}}{}
\stackunder[2pt]{\includegraphics[scale=0.135]{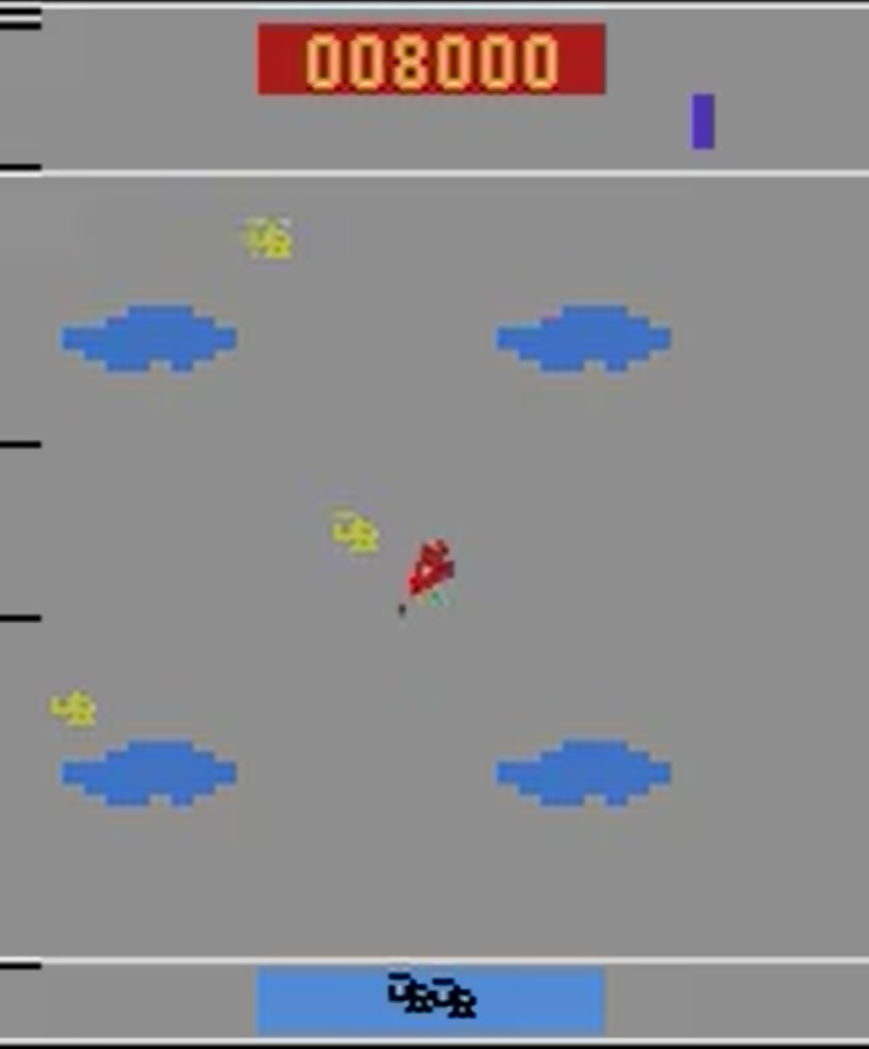}}{}
\stackunder[2pt]{\includegraphics[scale=0.135]{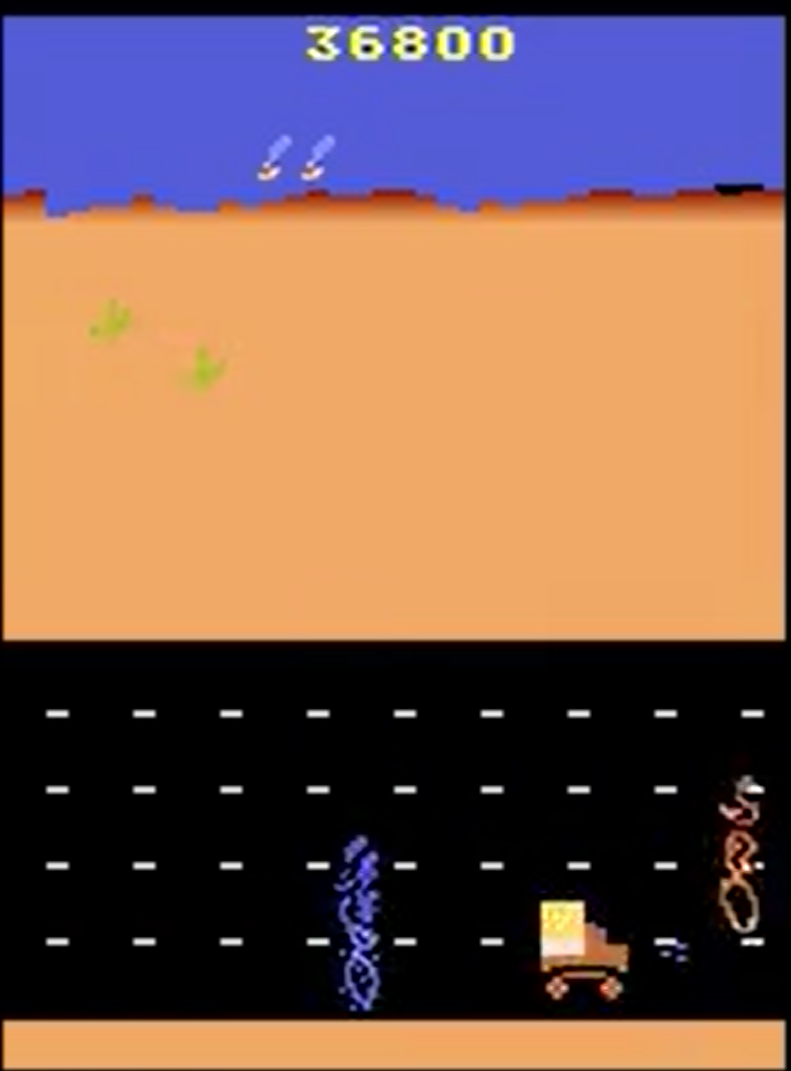}}{}
\stackunder[2pt]{\includegraphics[scale=0.175]{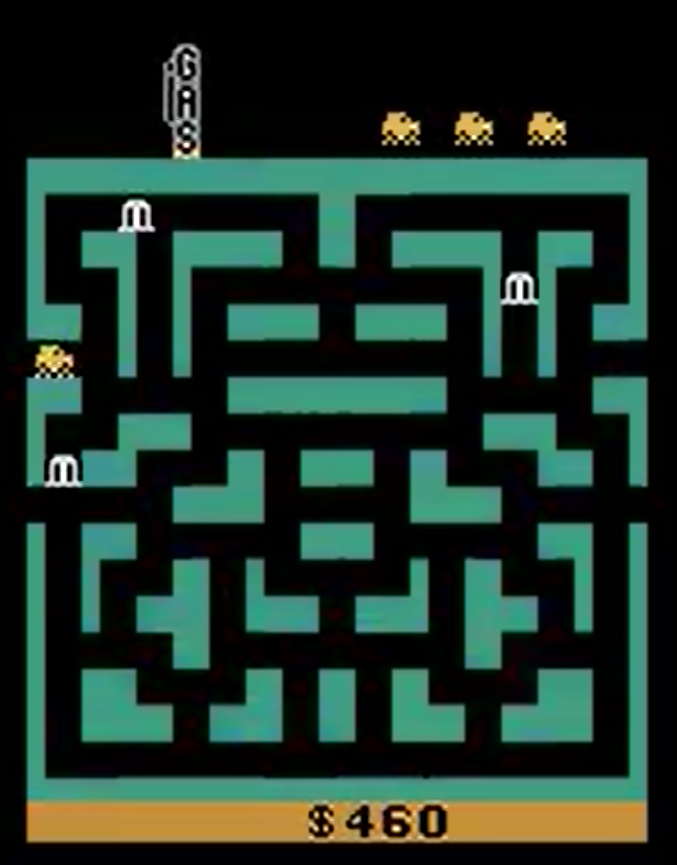}}{}\\
\stackunder[2pt]{\includegraphics[scale=0.15]{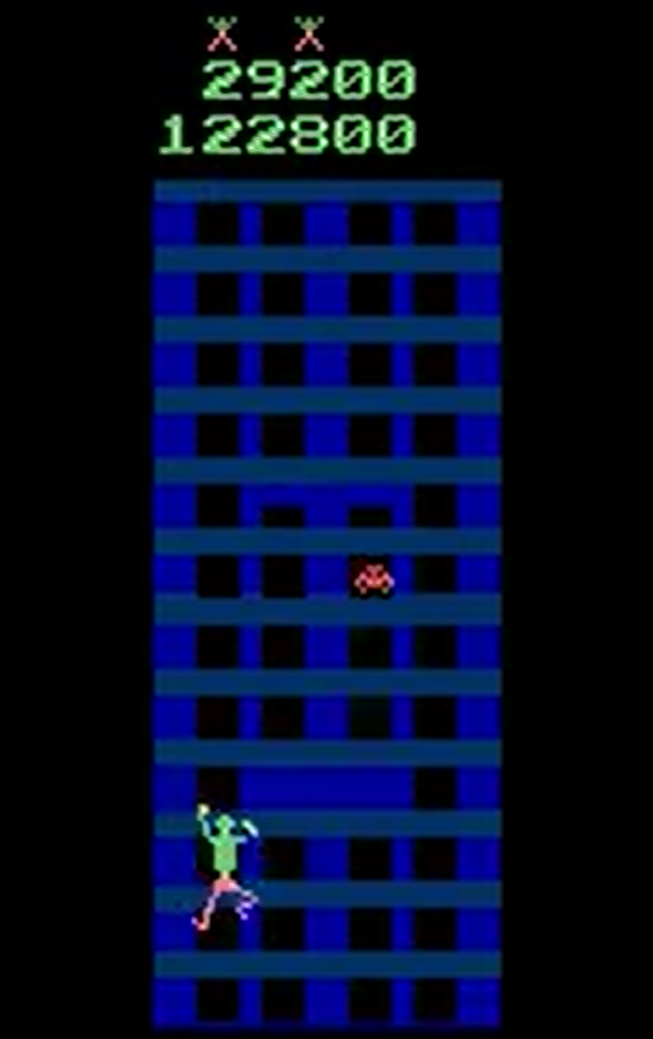}}{}
\stackunder[2pt]{\includegraphics[scale=0.135]{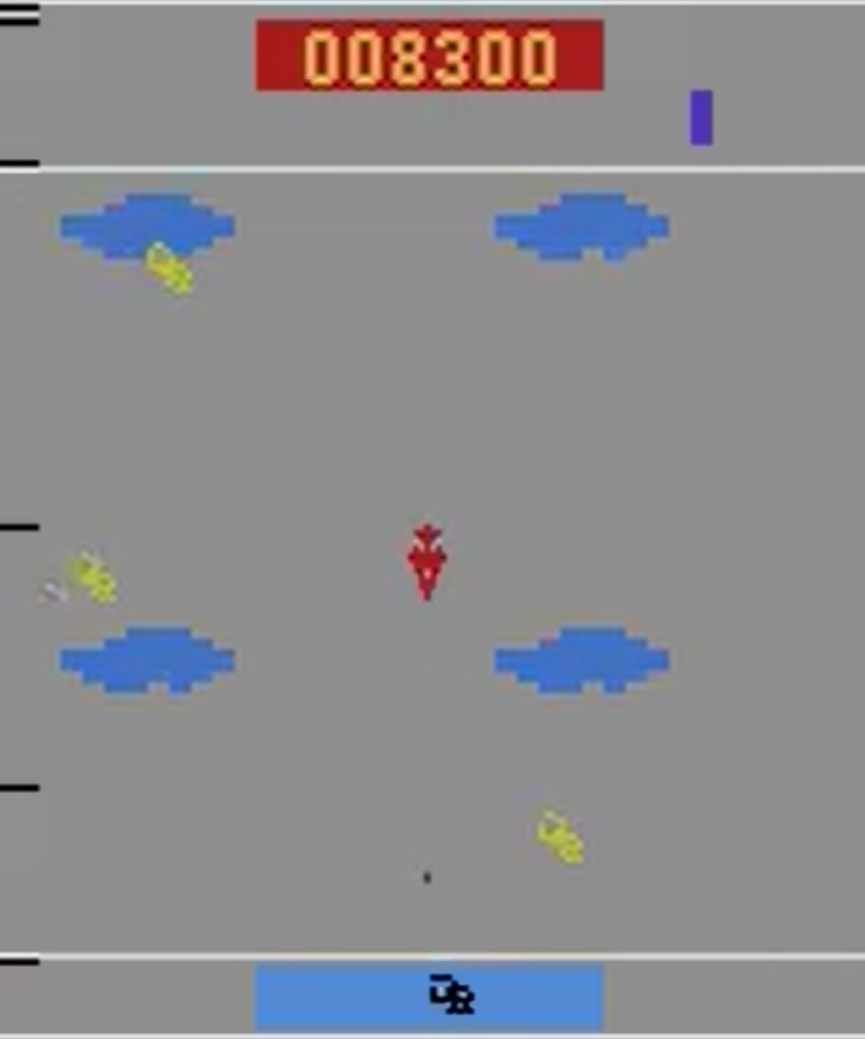}}{}
\stackunder[2pt]{\includegraphics[scale=0.135]{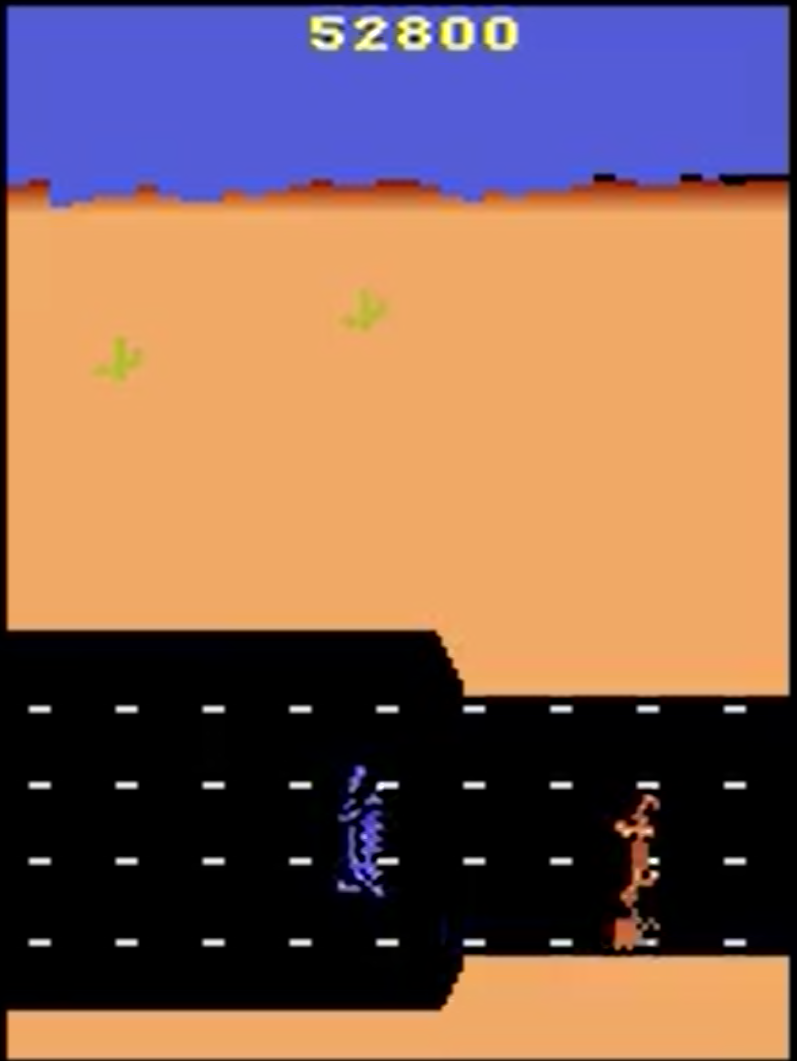}}{}
\stackunder[2pt]{\includegraphics[scale=0.175]{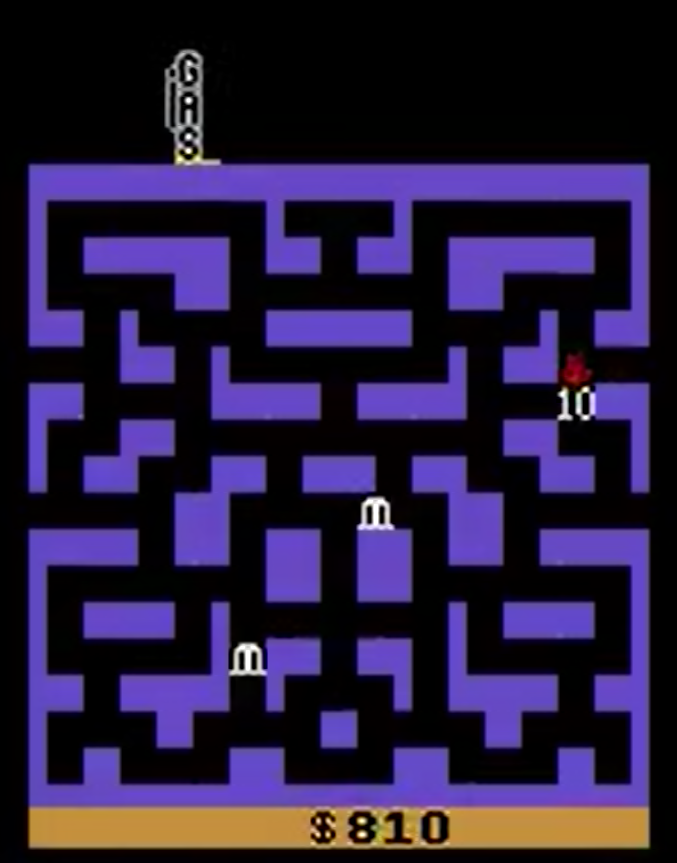}}{}
\end{center}
\caption{Markov Decision Processes from Arcade Learning Environment \citet{bell13}. Columns represent the rollout of states. Left: CrazyClimber. Middle Left: TimePilot. Middle Right: RoadRunner. Right: BankHeist.}
\label{environments}
\end{figure}

\section{Architectural Differences}
\label{para}

In this section we provide results for different DDQN architectures for the episode independent random state setting $\mathcal{A}_{\textrm{e}}^{\textrm{random}}$ and the environment independent random state setting $\mathcal{A}_{\mathcal{M}}^{\textrm{random}}$. Our aim is to investigate if architectural differences between different Double Deep Q-Networks yields to the creation of a different set of non-robust features. We have found that different architectures (e.g. Prior DDQN, Duel DDQN, Prior Duel DDQN) share the same non-robust features. In the experiments in this section the perturbation is computed in the DDQN Prior Duel architecture, and then added to the deep reinforcement learning policy's observation system trained with DDQN, DDQN Prior, DDQN Duel and DDQN Prior Duel.

\begin{table}[h!]
\caption{Impacts of the different settings within the proposed framework where the perturbation is computed in the DDQN Prior Duel architecture and added to the agent's observation system trained with DDQN, DDQN Prior, DDQN Duel and DDQN Prior Duel for the Pong environment.}
\label{arch}
\centering
\scalebox{1}{
\begin{tabular}{lccccccr}
\toprule
Architectures  & $\mathcal{A}^{\textrm{Gaussian}}$
				 & $\mathcal{A}_{\textrm{e}}^{\textrm{random}}$
				 &  $\mathcal{A}_{\mathcal{M}}^{\textrm{random}}$   \\
\midrule
DDQN                & 0.002     &  1.0           & 0.995     \\
DDQN Prior		      & 0.075     &  1.0            & 0.985     \\
DDQN Duel 		      & 0.036     &  1.0            & 0.966      \\
DDQN Prior-Duel     & 0.041     &  1.0           &  0.981     \\
\bottomrule
\end{tabular}
}
\end{table}

\end{document}

%% file: math_commands.tex
\usepackage{amsmath,amsfonts,bm,amsthm}

\newtheorem{theorem}{Theorem}[section]
\newtheorem*{theorem*}{Theorem}

\newtheorem{proposition}[theorem]{Proposition}
\newtheorem*{proposition*}{Proposition}

\newtheorem*{lemma*}{Lemma}

\theoremstyle{definition}
\newtheorem{definition}[theorem]{Definition}
\newtheorem*{definition*}{Definition}

\def\eqref#1{equation~\ref{#1}}

\def\1{\bm{1}}

\def\eps{{\epsilon}}

\def\vg{{\bm{g}}}

\def\vr{{\bm{r}}}

\def\vv{{\bm{v}}}
\def\vw{{\bm{w}}}

\DeclareMathAlphabet{\mathsfit}{\encodingdefault}{\sfdefault}{m}{sl}
\SetMathAlphabet{\mathsfit}{bold}{\encodingdefault}{\sfdefault}{bx}{n}

\def\gN{{\mathcal{N}}}

\def\sA{{\mathbb{A}}}

\newcommand{\E}{\mathbb{E}}

\newcommand{\R}{\mathbb{R}}

\DeclareMathOperator*{\argmax}{arg\,max}